\documentclass[12pt,a4paper]{article}

\usepackage{geometry}
\usepackage{amsmath}
\usepackage{cases}
\usepackage{amssymb,mathrsfs,dsfont}
\usepackage{enumerate}
\usepackage{enumitem}
\usepackage{tikz}
\usetikzlibrary{arrows}
\usepackage{mathtools}
\usepackage{float}
\usepackage{scrextend}
\usepackage{graphicx}
\usepackage{caption}
\usepackage{subcaption}
\usepackage{tikz,pgfplots}
\usepackage{graphicx}
\usepackage[utf8]{inputenc}

\usepackage{tikz}
\usepackage{dirtytalk}
\usetikzlibrary{positioning}
\usetikzlibrary{arrows,arrows.meta}
\tikzstyle{block}=[draw opacity=0.7,line width=1.4cm]
\usetikzlibrary{positioning,shapes,shadows,arrows,scopes,decorations}
\usetikzlibrary{matrix,chains,shapes.geometric,shapes.arrows,automata,chains,er,fit}

\tikzstyle{comment}=[rectangle, draw=black, fill=red!50!black, rounded corners, drop shadow,
anchor=west, text=white]
\usetikzlibrary{shapes.callouts}
\usetikzlibrary{mindmap}
\usetikzlibrary{calc,through,backgrounds}
\usepackage{setspace}

\def\m{\mathcal}

\def\unif{\mathsf{Unif}}
\def\fsm{\mathsf{FSM}}
\def\mc{\mathsf{MC}}
\def\sc{\mathsf{SC}}
\def\dfsm{\mathsf{DFSM}}

\def\dtv{d_{\mathsf{TV}}}
\def\Bern{\mathsf{Bern}}
\DeclareMathOperator{\Prob}{\mathbf{P}}
\newcommand{\veps}{\varepsilon}
\DeclareMathOperator{\E}{\mathbb{E}}
\DeclareMathOperator{\Pe}{\mathsf{P_e}}
\DeclareMathOperator{\Pstar}{\mathsf{P^*_e}}
\DeclareMathOperator{\ind}{\mathds{1}}

\DeclareMathOperator{\polylog}{polylog}

\DeclareMathOperator{\Var}{\mathsf{Var}}

\DeclareMathOperator{\isit}{\mathsf{RUNS}}

\title{Statistical Inference with Limited Memory: A Survey}
\author{Tomer Berg, Or Ordentlich, Ofer~Shayevitz}
\date{}

\begin{document}
\maketitle

\begin{abstract}%
The problem of statistical inference in its various forms has been the subject of decades-long extensive research. Most of the effort has been focused on characterizing the behavior as a function of the number of available samples, with far less attention given to the effect of memory limitations on performance. Recently, this latter topic has drawn much interest in the engineering and computer science literature. In this survey paper, we attempt to review the state-of-the-art of statistical inference under memory constraints in several canonical problems, including hypothesis testing, parameter estimation, and distribution property testing/estimation. We discuss the main results in this developing field, and by identifying recurrent themes, we extract some fundamental building blocks for algorithmic construction, as well as useful techniques for lower bound derivations.

\end{abstract}

\section{Introduction}

Statistical inference is one of the most prominent pillars of modern science and engineering. Generally speaking, a statistical inference problem is the task of reliably estimating some property of interest from random samples, obtained from a partially or completely unknown distribution. Such properties can include, for example, the mean of the distribution, its higher moments, whether the distribution belongs to a certain class (e.g., is uniform or Gaussian) or is far from that class, the entropy of the distribution, or even the distribution itself. These types of problems arise constantly in many research areas, both theoretical and practical, and across multiple disciplines in engineering, computer science, physics, economics, biology, and many other fields and subfields within. 

The extremely broad literature on statistical inference has been concerned almost exclusively with an idealized setting, in which no practical constraints are posed on the inference algorithm or the data collection process. Specifically, two implicit postulates underlying this classical setup are that inference is completely \textit{centralized}, namely that all the samples are available to a single agent who carries out the inference procedure, and that this agent has essentially \textit{unlimited computational power}, hence can implement the optimal decision rule. These crucial assumptions, however, often fail to hold in real world settings, where data is either collected in a distributed fashion by multiple remotely-located agents who are subject to stringent communication constraints, or by a single agent who has limited computational capabilities (or both).\footnote{Another common constraint is privacy, and in particular local differential privacy, where the central server is not allowed to learn much about the samples of any individual agent. We refer the interested reader to a survey and a couple of state of the art works on this topic~\cite{dwork2008differential,duchi2013local,acharya2021inference}.} If the number of collected samples is very small relative to the computational power or the available communication bandwidth, then the computationally-limited / distributed setup can be easily reduced essentially to a computationally unlimited / centralized one with little overhead. However, this is not possible when either communication or computation / memory become a bottleneck. 

As a simple example, imagine a network of low-complexity remote sensors sampling their local environments and reporting back over a low-bandwidth channel to a
base station, whose goal is to monitor irregular behavior in the network. The amount of local data observed by each sensor might be quite large (e.g., video, internet traffic, etc.) and cannot be fully processed / locally stored, nor completely conveyed to the base-station in time to facilitate fast anomaly detection. To meet their goal, the sensors would therefore need to send a much shorter and partial description of their samples. What is the shortest description they can send that still allows the base station to achieve its desired level of minimax risk? What is the effect of limited computational power on the attainable risk? Can these limits be efficiently achieved, and if so, how?  This types of timely questions have gained much contemporary interest. 

In this survey, we focus on the computational aspect of such problems, and in particular on the limited memory aspect, where an inference algorithm is restricted to store only a small number of bits in memory. The study of learning and estimation under memory constraints has been initiated in the late 1960s by Cover and Hellman \cite{cover1969hypothesis,hellman1970learning} (with a precursor by Robbins~\cite{robbins1956sequential} on the two armed bandit problem), and remained an active research area for a few decades. It has then been largely abandoned, but has recently drawn much attention again, with many works addressing different aspects of inference under memory constraints in the last few years. Despite this ongoing effort, there are still gaps in our understanding of the effects that memory limitation has on inference accuracy in various problems, as well as its impact on the required sample size. In this survey, we attempt to provide a snapshot overview of this fast-evolving area, summarizing the main results in the field and the landscape around it. We also delineate a few high level ideas and techniques often used to obtain upper and lower bounds in this type of problems, point out some interesting phenomena and trade-offs, and discuss open problems. 

\textit{Outline}. In Section~\ref{sec:problem} we formally define the general inference setting, present the different types of inference problems we will encounter in the paper, and introduce the setting of inference under finite-state memory constraints. In Section~\ref{sec:algs}, we discuss some assumptions on finite-state algorithms, including time-invariance and randomization. Section~\ref{sec:related} is a short digression from the main theme of the paper, where we briefly survey the closely-related problems of inference in the data stream model, and inference under communication constraints. Section~\ref{sec:meth} introduces the main methodologies used in deriving lower bounds and upper bounds in the memory constrained setting. Sections~\ref{sec:hyp} thru~\ref{sec:prop_est} give a thorough overview of the landscape and current status of knowledge in memory constrained hypothesis testing, parameter estimation, distribution testing and distribution property estimation problems. Finally, Section \ref{sec:open} addresses open problems and future research directions.

\section{Problem Setting and Definitions}\label{sec:problem}

\subsection{The Inference Problem}

In classical statistical inference, one is often concerned with the following prototypical problem. Nature picks some unknown probability distribution $P$ from a given family $\m{F}$ of possible distributions, and an agent wants to estimate some property $\pi (P)\in \m{Y}$ of the distribution. The property space $\m{Y}$ can be taken in most cases to be some subset of $\mathbb{R}^d$ for some $d$. To that end, the agent observes $n$ i.i.d samples $X^n=(X_1,\ldots,X_n)$ from $P$ and computes an estimator $\hat{\pi}:\m{X}^n\to \m{Y}$ for $\pi(P)$. The quality of estimation is measured w.r.t some loss function $\ell:\m{Y}\times\m{Y}\to \mathbb{R}_+$, which gives rise to the notion of the \textit{expected risk}, defined as
\begin{align}
    R_n(P,\hat{\pi})\triangleq \E [\ell(\pi(P),\hat{\pi}(X^n))].
\end{align}
The agent’s goal is to find an estimator $\hat{\pi}$ that is optimal in some sense. Various notions of optimality appear in the literature, including those based on the \textit{Bayesian paradigm} and on the \textit{unbiased estimation} approach. Here, we focus on the more robust \textit{minimax paradigm}, since it makes no assumptions on the
behavior of the data or the structure of the estimator. Under this paradigm, the agent chooses the estimator
$\pi^*_n$ that guarantees the minimal possible risk in the worst case, i.e., attaining  the so-called \textit{minimax risk} of the problem, defined as
\begin{align}\label{eq:general_minimax_risk}
    R_n^*\triangleq \inf_{\hat{\pi}}\sup_{P\in \m{F}}R_n(P,\hat{\pi}).
\end{align}
Most of the works in the machine learning literature have characterized the statistical hardness of the inference problem by the \textit{minimax sample complexity}, which is the minimal number of samples $n$ for which $R_n^*\leq \delta$ can be guaranteed uniformly for all $P\in \m{F}$, for some fixed target risk $\delta > 0$, i.e.,
\begin{align}
   \mathsf{SC}(\delta)=\min \{n:R_n^*\leq \delta\}. 
\end{align}
The theory of minimax inference is well understood in many statistical inference settings, and here is a short, non exhaustive list~\cite{tsybakov1997nonparametric,devroye2001combinatorial,lehmann2006theory,goldreich2010property,cai2011testing,jiao2015minimax,wu2016minimax}.

\subsection{Problem Types}\label{sec:prob}
Let us briefly recall some inference problems that have been extensively studied in the traditional (computationally-unbounded) setting. We later revisit these problems under our computation restricted paradigms. 

\subsubsection{Simple Hypothesis Testing}
In this problem, the samples are drawn from a member in a discrete family of distributions, $\m{F}=\{P_0,\ldots,P_{d-1}\}$ and we are interested in guessing the correct underlying distribution $\pi(P_i)=i$. Some common problems are ascertaining whether the samples come from a particular population (say, infected/healthy), or identifying whether they are generated by a signal suffused in noise in contrast to originating from pure noise (the RADAR problem). The frequently used loss function in this case is the $0/1$ loss, $\ell (\pi,\hat{\pi})=\ind(\pi \neq \hat{\pi})$, which gives rise to the expected risk
\begin{align}
    \E[\ell (\pi,\hat{\pi})]=\E[\ind(\pi \neq \hat{\pi})]=\Pr(\pi \neq \hat{\pi}),
\end{align}
which is precisely the error probability of the algorithm. The minimax risk is hence the smallest error probability that can be uniformly guaranteed for all hypotheses in $\m{F}$. A secondary line of work, mostly appearing in the information theory literature, is interested in the decay rate of $\Pr(\pi \neq \hat{\pi})$ as a function of the number of samples $n$, and, particularly, in the optimal \textit{error exponent} for Neyman–Pearson testing, which is the decay rate of the type II error probability subject to the constraint that the type I error probability is at most some specified constant. For a more symmetric approach, one may consider the Bayesian setting, in which prior probabilities are assigned to both hypotheses and we want to minimize the average error probability, or the minimax setting, in which we want to minimize the maximal error probability. The best achievable error exponent in both cases is the \textit{Chernoff information} or \textit{Chernoff Divergence}~\cite{cover2012elements,neyman1933ix,tuncel2005error}.

\subsubsection{Parameter Estimation}
In this setup, the family of distributions is parameterized $\m{F}=\{P_{\theta}\}_{\theta \in \mathbb{R}^d}$, and we would like to approximate the value of the parameter itself, i.e., $\pi(P_{\theta})=\theta$. Some common problems are estimating the bias of a coin and location estimation in sensor networks. The loss function of choice here is the squared error loss, $\ell(\hat{\theta},\theta)=(\hat{\theta}-\theta)^2$, which brings rise to the expected risk
\begin{align}
   \E[\ell(\hat{\theta},\theta))] = \E[(\hat{\theta}-\theta)^2],
\end{align}
known as the \textit{quadratic risk}. Thus we are looking for the \textit{minimax quadratic risk}, also referred to as minimum mean squared error.

\subsubsection{Composite Hypothesis Testing}
In this more general problem the family of distributions is a finite disjoint union of sub-families $\m{F}=\cup_{i=1}^{d-1}\m{F}_i$, and we want to find out whether the underlying distribution belongs to one of the sub-families. The loss is again the $0/1$ loss, thus the expected risk is the minimax error probability. Of a particular interest to us is the \textit{distribution property testing} variant, on which we elaborate below.

\subsubsection{Distribution Property Testing} 
This is a special case of composite hypothesis testing, where we are interested in deciding whether the underlying distribution has a certain property or is $\veps$-far in total variation from having this property. Common problem are testing whether the distribution is uniform, sparse, or has entropy below some threshold value. There are two composite hypotheses in this problem, where $\m{F}_0$ contains all the distributions with the property, and $\m{F}_1$ contains all the distributions $P$  that satisfy 
\begin{align*}
   \inf_{Q\in \m{F}_0}\dtv (P,Q)> \veps, 
\end{align*}
where $\dtv (P,Q)$ is the total variation distance between $P$ and $Q$, which is defined as half the $\ell_1$ distance, i.e., $\dtv (P,Q)=\frac{\lVert P-Q \rVert_1}{2}$.

\subsubsection{Distribution Property Estimation}
The family of distributions is again parameterized by $\m{F}=\{P_{\theta}\}_{\theta \in \mathbb{R}^d}$, but here we are trying to learn some ``nice'' property of the distribution induced by $\theta$, e.g., some simple scalar function $h(P_{\theta})$ instead of $\theta$ itself. Common problems are approximating the Shannon entropy of a distribution, or approximating the support size (also known as the distinct element problem). The loss function is often taken to be the $\ell_1$ distance, which gives rise to the \textit{$\ell_1$ risk}, 
\begin{align}
  \E[\ell(\hat{h},h)]=\E|\hat{h}-h|.
\end{align}
In the machine learning literature works dealing with this problem, the loss function is sometimes taken to be the $0-1$ loss w.r.t. an $\veps$-error event in $\ell_1$, defined as
\begin{align}
    \ell(\hat{h},h)=\ind(|\hat{h}-h|>\veps),
\end{align}
where $\veps>0$ is a given design parameter. Namely, here the risk is the probability that our estimator is too far from the ground truth, i.e., $\Pe=\Pr(|\hat{h}-h|>\veps)$. This is also known as the Probably Approximately Correct (PAC) setting~\cite{valiant1984theory}. 

\subsection{Memory Constraints}
In the limited-memory framework, an agent observes an i.i.d sequence $X_1,X_2,\ldots \in \m{X}$, and at each time point is allowed to update a memory register, or, equivalently, a memory state. Throughout the survey, we will use the memory states terminology to refer to a memory limited algorithm, as it is consistent with many of the earlier works and some of the contemporary ones. Specifically, let $\fsm(S, \mathcal{X}, \m{Y})$ denote the family of all \textit{finite-state machines} over the state space $[S]=\{1,2,\ldots,S\}$, with input space $\m{X}$ and output space $\m{Y}$. We will usually write $\fsm(S)$ for short, leaving the input and output spaces implicit. A finite-state machine, or finite-state algorithm, is a pair $(f,g)$ of a \textit{state-transition function} $f:[S] \times \m{X}\rightarrow [S]$, and a \textit{state-decision function} $g:[S]\to \m{Y}$. Without loss of generality, we can assume the finite-state machine is initialized to state $1$ at time $n=0$. When fed with the input sequence $X_1,X_2,\ldots$, the state of the machine evolves as follows:
\begin{align}
T_0&=1\\
   T_n&=f(T_{n-1},X_n).\label{eq:trans}
\end{align}
Correspondingly, the state-decision at time $n$, which in our problem yields the estimator $\hat\pi$ for $\pi$ at time $n$, is given by 
\begin{align}
    \hat\pi(X^n) = g(T_n).
\end{align}
With a slight abuse of notation, we will write $\hat\pi\in \fsm(S)$ to mean that our estimator is of the above structure, i.e., is obtained as the state-decision of some finite-state machine with $S$ states. We will refer to $\hat\pi$ as finite-state estimation algorithm, or sometimes for brevity, simply a finite-state estimator.


We emphasize that the state-transition and state-decision rules $(f,g)$ we consider here are \textit{time-invariant}, i.e., they are fixed and do not depend on the time parameter $n$. It is also possible to consider time-varying machines, namely to let $(f_n,g_n)$ depend on $n$. We will discuss the time-varying case in Subsection~\ref{sec:time_var}, and explain why the restriction to time-invariant machines is reasonable. 

We include \textit{randomized} machines in our $\fsm$ family, i.e., machines for which the transition function $f$ or the decision function $g$ depend on some exogenous randomness, independent of the samples. We will elaborate more on randomized algorithms in Subsection~\ref{sec:rand}, where we highlight some of the advantages randomized algorithms possess over deterministic algorithms.  

In this survey, we are mainly interested in the minimax risk $R_n^{*}(S)$ of the memory-constrained inference problem. This risk is similar to the unconstrained minimax risk~\eqref{eq:general_minimax_risk}, with the additional structural constraint requiring the algorithm producing the estimator $\hat\pi$ to be finite-state. Formally: 
\begin{align}
R_n^{*}(S)=\inf _{\hat\pi\in \fsm(S)}\sup_{P \in \m{F}} \sup_{k\geq n} R_k(P,\hat{\pi}).  
\end{align}
Note that we are maximizing the risk also over the number of samples $k\geq n$. When there are no memory constraints, this additional maximization is superfluous, since the minimax risk can only improve with the number of samples. But with limited memory, it is possible to design finite-state algorithms that are tailored to each $n$ separately. Hence, the maximization over $k\geq n$ guarantees that the machine is not ``aware'' of the time horizon, beyond being at least $n$ samples. 

In many of the works covered in this survey, one is interested in the asymptotic behaviour of a finite-state algorithm, that is, in the risk attained with fixed memory in the limit of many samples $n\rightarrow \infty$. To that end, we define the \textit{asymptotic risk} of an estimator to be
\begin{align}
   R(P,\hat{\pi}) = \limsup_{n\rightarrow\infty}   R_n(P,\hat{\pi}), \label{eq:infinity_risk} 
\end{align}
Note that using $\limsup$ is required in order to cover FSM algorithms that have periodic components, for which the regular limit does not exist. In some of the works covered in the survey, this issue is circumvented using an alternative definition of the asymptotic risk as the asymptotic average of the instantaneous risks:
\begin{align}
   \bar{R}(P,\hat{\pi})=\lim_{n\rightarrow\infty}\frac{1}{n}\sum_{t=1}^n R_t(P,\hat{\pi}) .
\end{align}
We note that for our purposes, there is no restriction in using $R(P,\hat{\pi})$ instead of $\bar{R}(P,\hat{\pi})$. This follows since on the one hand, $\bar{R}(P,\hat{\pi})\leq R(P,\hat{\pi})$, so any lower bound on $\bar{R}(P,\hat{\pi})$ is also a lower bound on $R(P,\hat{\pi})$, and on the other hand, all algorithms appearing in this survey have no periodic components, and for such algorithms both definitions coincide. 

We can now define the \textit{asymptotic minimax risk} to be
\begin{align}
  R^{*}(S)=\inf _{\hat\pi\in \fsm(S)}\sup_{P \in \m{F}} R(P,\hat{\pi}). \label{eq:minimax_infinity}
\end{align}
While in general the algorithm $\hat\pi$ might be randomized, in some cases we will be interested in the performance of deterministic algorithms, i.e., algorithms that do not have access to external randomness (as we elaborate on in section~\ref{sec:rand}). We thus define an additional notion of minimax risk, which we will refer to as the \textit{asymptotic deterministic risk}, defined as
\begin{align}
    R_{\mathsf{det}}^*(S) = \inf _{\hat\pi\in \dfsm(S)}\sup_{P \in \m{F}} R(P,\hat{\pi}), \label{eq:det_infinity}  
\end{align}
where $\dfsm(S)$ is the family of finite-state \textit{deterministic} estimation algorithms.
Let $\delta^*\geq 0$ be the smallest obtainable minimax risk, i.e.,
\begin{align*}
  \delta^*=\min \{\delta: \exists S,n \in \mathbb{N}^+,R_n^*(S)\leq \delta\}. 
\end{align*}
For any $\delta^*\leq \delta$, we can define the \textit{sample-memory} curve at level $\delta$, which is a curve on the $(n,S)$ axis, that represents the minimal number of samples $n$ and the minimal number of states $S$ needed to attain a minimax risk of $\delta$. The curve can be characterized by the function
\begin{align}
S^*(n,\delta)=\min \{S:R_n^{*}(S)\leq \delta),
\end{align}
which gives the minimal number of states $S$ required for the minimax risk to be at most $\delta$, or, equivalently, by
\begin{align}
n^*(S,\delta)=\min \{n:R_n^{*}(S)\leq \delta), 
\end{align}
which gives the minimal number of samples $n$ required for the minimax risk to be at most $\delta$. Note that these two functions are in an inverse relation. The sample-memory curve at level $\delta$ naturally yields two asymptotic quantities: First, the classical notion of sample complexity is obtained as the minimal number of samples required to attain a risk $\delta$ in the limit of infinite (or large enough) memory: 
\begin{align}
  \mathsf{SC}(\delta)=\lim_{S\rightarrow \infty} n^*(S,\delta), 
\end{align}
and second, the notion of \textit{memory complexity}, which we define here as the smallest number of states required to attain a risk $\delta$, in the limit of an arbitrarily large number of samples: 
\begin{align}
  \mathsf{MC}(\delta)=\lim_{n\rightarrow \infty}S^*(n,\delta).  \label{eq:mc}
\end{align}

Note that we define the memory complexity as the number of \textit{states} the algorithm uses, rather than the number of \textit{bits}; this is just a matter of convenience, since often writing results using the number of states is more natural and readable, having the same ``units'' as number of samples. We will nevertheless occasionally use bits for memory size, in which case we will use the term \textit{memory complexity in bits}.  Note that the memory complexity in bits never exceeds $\mathsf{SC}(\delta)\cdot \log|\m{X}|$, as this is the number of bits needed to trivially achieve a risk of $\delta$ by naively saving the first $\mathsf{SC}(\delta)$ samples.

Characterizing the sample-memory curve in general is a very difficult task, and we are currently unaware of any non-trivial problem for which the sample-memory curve at level $\delta$ is completely known. Indeed, most of the classical research was focused on only characterizing the two extreme points of $\mathsf{SC}(\delta)$ and $\mathsf{MC}(\delta)$, and only in recent years have other points on the curve been addressed (as we expand upon in later sections). 

\section{Algorithm Types}\label{sec:algs}
In this section, we discuss different assumptions that can be made on finite memory algorithms. First, we consider the dependence of the algorithms on time, and argue that it makes sense to restrict the discussion to time-invariant algorithms. We then consider randomized algorithms, i.e. ones that have access to an external source of randomness, and contrast them with deterministic algorithms, where the only randomness comes from the samples.

\subsection{Time-Varying vs. Time-Invariant}\label{sec:time_var}
A finite-memory estimation algorithm $\hat{\pi}$ is said to be \textit{time-varying}, if either the state-transition function $f$, or the state-decision function $g$ (or both) are allowed to depend the time index $n$. Otherwise, namely if both $f$ and $g$ are fixed for all $n$, the algorithm is called time-invariant.
We note that any time-varying algorithm can be trivially reduced to a time-invariant one with an $O(\log{n})$ bits overhead, if the total number of available samples $n$ is finite, by simply keeping a clock.\footnote{We do not account here for the complexity of describing the state transition function $f$. In time-varying algorithms, this description may grow as a function of the number of observed samples.} Thus, in our framework, a time-varying algorithm implicitly means that we assume an external clock is provided ``for free'', i.e., without accounting for the memory required to store it. Since what we care about is the memory budget, it seems less preferable to make this kind of an assumption.  Moreover, when the number of samples is unbounded, this assumption can lead to absurd results. For example, one can solve the two-armed Bandit problem with only $2$ states~\cite{cover1968note}, and the binary hypothesis testing problem using just $3$ states and with arbitrarily small error (!) \cite{cover1969hypothesis,koplowitz1975necessary}. Finally, as we shall see, some optimal time-invariant algorithms inherently do not keep a clock at all. For all these reasons, in this survey we will restrict our attention to time-invariant algorithms.

\subsection{Randomized vs. Deterministic}\label{sec:rand}
A randomized algorithm is an algorithm whose state-update and/or state-decision functions depend not only on the samples, but also on some external randomness independent of the samples. A natural question is in what cases randomization even helps? Note that, trivially, there is no loss of generality in assuming that the state-decision function is deterministic; this follows simply since the risk incurred by a random decision rule is some average over risks obtained by deterministic ones, hence one can always pick the risk-minimizing deterministic rule. 

Importantly however, this argument \textit{does not} apply to the state-transition function, since picking the particular randomness realization that minimizes the risk over the choice of transition functions will generally result in a time-varying machine, which is not in $\fsm(S)$. Indeed, for time-varying algorithms it is sufficient to consider deterministic algorithms as the optimal time-varying rule is deterministic (see~\cite{cover1976optimal}). For time-invariant algorithms, we will see in this survey that there are certain problems where randomization improves the limiting behaviour, characterized by the asymptotic notation $O(\cdot)$, of a finite-state machine, e.g., binary hypothesis testing under the $0-1$ loss, and others where it does not, e.g., bias estimation under the squared loss measure (as we show in the sequel).

Now, while allowing the use of randomization in the estimation procedure may certainly help, this resource has a cost. Even if one has access to unlimited randomness (which is the case, for example, when observing an unlimited number of samples, since randomness can be extracted from the i.i.d. sequence $X_1,X_2,\ldots$), storing this randomness places a toll on one's memory budget, which is taken into account in the deterministic setup. In general, we would seek upper bounds on deterministic algorithm as they are more restrictive than randomized ones, and lower bounds for randomized algorithms, which then carry on to deterministic algorithms due to the reasons mentioned above. 

\section{Related Problems}\label{sec:related}

We briefly comment on two models related to inference from i.i.d. samples under memory constraints. The first is the data stream model, in which no statistical assumptions are made on the data, and the algorithms are designed for the worst case. The data stream model is hence fundamentally more difficult than the inference setting addressed in this paper. The second setup we consider in this section is the communication-constrained setup. Here, samples are collected by remotely located agents, and the inference bottleneck is imposed by a limited communication budget between the agents. There are several models that capture the limited communication links, some of which are easier then the memory-constrained setup. One such example is the \emph{blackboard model}, where each agent has access to all the messages sent by previous agents, and each message can take $S$ values.  This model is easier than our memory constrained model, in which each agent can choose its message using a finite-state machine whose input is the current observation and the message sent by the previous agent alone. Please note that this section is optional and can be skipped. 

\subsection{The Data Stream Model}\label{sec:stream}

The setting that we cover in this survey is closely-related to the \textit{data stream model} in theoretical computer science, yet there are principal differences between the two. For clarity, in this section we briefly highlight the similarities and pinpoint the differences between the models.

The data stream model characterizes algorithms with small memory size that make passes over the input in order; any item not explicitly stored is inaccessible to the algorithm in
the same pass. In many cases the number of passes is limited to one. The figure of merit is taken to be the memory complexity in bits, usually referred to in these works as the \textit{space complexity} of a problem. 
The memory complexity in bits of many different streaming algorithms have been extensively studied within theoretical computer science, and some well known examples include estimation of frequency moments of the data stream~\cite{alon1999space,indyk2005optimal}, estimation of Shannon and Rényi entropy of the empirical distribution of the data
stream~\cite{lall2006data,chakrabarti2006estimating,chakrabarti2010near}, estimation of heavy hitters~\cite{charikar2002finding,cormode2005improved,metwally2005efficient}, and estimation of distinct elements~\cite{flajolet1985approximate,durand2003loglog,indyk2003tight,flajolet2007hyperloglog,kane2010optimal}.
Though some exceptions exist~\cite{guha2009sublinear,chakrabarti2008tight}, the vast majority of works are concerned with worst case inputs (i.e., adversarially generated data), and thus lower bounds for the data stream model do not generally apply in the statistical learning setup, where the data is often assumed to be drawn i.i.d. from some unknown underlying distribution.

In general, the data stream model is substantially more difficult than the stochastic setup. To demonstrate this point, we provide a couple of examples for problems where the memory complexity required for evaluating some function of the empirical distribution of a worst case input tends to be significantly larger than the memory complexity for evaluating the same function of the underlying distribution in a stochastic setup. Consider, for example, the problem of entropy estimation of a data stream. To provide an $\veps$ multiplicative approximation of the empirical entropy of a stream of $n$ values in a single pass, one must use at least $\Omega( \veps^{-2}/ \log^2(\veps))$ memory bits, according to the work of Chakrabarti et al.~\cite{chakrabarti2010near} (the lower bound is proved for the assignment $\veps^{-1}=3\sqrt{n}(\log n +1/2)$). In contrast, when approximating the entropy of a distribution over alphabet $\m{X}=[k]$ with the same multiplicative factor, only $O(\log (k/\veps))$ memory bits are needed (see~\cite{acharya2019estimating,aliakbarpour2022estimation}). This entails an \textit{exponential} increase in the number of necessary memory bits w.r.t the dependence on $\veps$. 

Another example follows from the problem of approximating the second frequency moment of a sequence, also known as the \textit{repeat rate} or
\textit{Gini’s Index of homogeneity}. Woodruff~\cite{woodruff2004optimal} has shown that any algorithm that $\veps$ approximates the second moment requires at least $\Omega(1/\veps^2)$ memory bits, for any $\veps=\Omega(1/\sqrt{k})$, where $k$ is the alphabet size. The statistical variant of the problem is approximating the collision probability of a distribution over alphabet $\m{X}=[k]$. A simple algorithm that performs this is the following: define a collision machine
to be a machine that, at each time point, outputs $1$ if the current sample is the same as the preceding sample and $0$ otherwise. This can be implemented using $\log k$ bits, since we are only storing the preceding sample, and it results in a Bernoulli process with parameter $p\geq 1/k$. As we show elsewhere in the survey, this Bernoulli parameter can be approximated using $O(\log (k/\veps))$ bits, resulting in an overall $O(\log (k/\veps))$ memory bits. Thus, again, we have an exponential increase with respect to the loss parameter $\veps$. 

There are, however, some interesting rare cases in which the adversarial setting does not cost more memory than the stochastic one. A specific example arises from the field of universal prediction with finite-state machines. In a surprising result, Meron and Feder~\cite{meron2004finite} showed that a $K$-state randomized finite-memory predictor for
worst case sequences achieves a $\asymp 1/K$ redundancy w.r.t to the least square loss, which is the same redundancy it incurs in the stochastic Bernoulli setting. Another example stems from the problem of distinct elements approximation: It is known, due to the works of Indyk and Woodruff~\cite{indyk2003tight} and Kane et al.~\cite{kane2010optimal}, that the memory complexity in bits of the distinct elements problem is $\asymp 1/\veps^2+\log n$, when the input length is $n$ and $\veps$ is the approximation parameter. Woodruff~\cite{woodruff2009average} initiated the study of average-case complexity of the distinct elements problem under certain distributions and, in particular, considered a stochastic setup in which we observe an $n$-length random sequence drawn from an alphabet size $k$, where each element is drawn uniformly and independently
from an unknown subset of some unknown size $d$. The main contribution of the paper is to show that if $n, d \asymp 1/\veps^2$, then the problem requires $\Omega(1/\veps^2)$
memory bits. Thus, even in the random data model, distinct elements estimation can be as memory-complex as its data stream variant. 

\subsection{Inference Under Communication Constraints}
The topic of inference under communication constraints is not covered in this survey. Nevertheless, as it is closely related to inference under memory constraints (as we will later see), we provide a short and non-exhaustive overview of this problem. The study of statistical inference under communication constrains has originally been considered in the information theoretic literature. Parameter estimation under communication constraints was initially studied by Zhang and Berger~\cite{zhang1988estimation}, who provided a rate-distortion upper bound on the quadratic error in distributed estimation of a scalar parameter using one-way communication under a rate constraint of bits per sample. There has been much follow-up work on this type of problems, especially in the discrete alphabet case, see, in particular~\cite{amari1989fisher,ahlswede1990minimax,amari1995parameter,amari1998statistical,amari2011optimal} and references therein. The closely related problem of distributed hypothesis testing against independence under communication constraints was originally treated by Ahlswede and Csis\'zar~\cite{ahlswede1986hypothesis}, who provided an exact
asymptotic characterization of the optimal tradeoff between the rate of one-way protocols, and the Type I error exponent attained under a vanishing Type II error probability. This result has been extended in many interesting ways, see, e.g.,~\cite{amari1989statistical,shalaby1992multiterminal,wigger2016testing,watanabe2017neyman,weinberger2019reliability,sreekumar2019distributed,salehkalaibar2020distributed}. Specifically, a generalization to the interactive setup with a finite number of rounds was addressed in~\cite{xiang2013interactive}, and the associated one-way communication complexity problem was recently studied via hypercontractivity in~\cite{sahasranand2018extra}.

In a different line of work~\cite{hadar2019communication,hadar2019distributed}, Hadar et al. assumed that the agents have access to an arbitrarily large number of local samples, and are only limited by the number of bits they can exchange. In this setup, it is assumed that the underlying distribution is unknown, and the goal is to estimate correlations in the minimax sense. They thus define the \textit{communication complexity} of the inference task to be the minimal number of bits required to guarantee that the minimax risk is below some constant, in the limit of large number of local samples. This definition is different from the definition of communication complexity as presented in the machine learning literature, which is the least possible number of bits needed to be interactively exchanged by Alice and Bob, such that they are able to compute some given function of their local inputs, either exactly or with high probability, see, e.g.,~\cite{yao1979some,kushilevitz1997communication,rao2020communication} . 

Finally, there has also been much contemporary interest in distributed inference problems with communication
constraints in a different context, where a finite number of i.i.d. samples from an unknown distribution are observed by multiple remotely located parties, who communicate with a data center under a communication
budget constraint (one-way or interactively), to obtain an estimate of some property of interest. In these type of problems, the samples observed by the
agents are from the same distribution and the regime of interest is typically where the dimension of the
problem is high compared to the number of local samples, see, e.g.,~\cite{zhang2013information,garg2014communication,acharya2020domain,braverman2016communication,xu2016information,jordan2018communication,han2018geometric,fischer2018distributed,acharya2019inference,acharya2020inference,acharya2021inference,pensia2022communication}. This problem bares close resemblance to remote source coding under Gaussian noise, known as the CEO problem, which have been extensively studied in the information theory literature, see, for example,~\cite{berger1996ceo,viswanathan1997quadratic,eswaran2019remote} and Chapter $33.11$ of Polyanskiy and Wu~\cite{polyanskiy2022information}.

\section{Methodologies}\label{sec:meth}
In this section, we provide an overview of common methodologies and ideas used in deriving performance bounds in finite-memory inference. For lower bounds, we describe the method of reduction to a communication complexity problem (which originated from works on the streaming model), the information theoretic approach, and the approach of Markov chain analysis. We then delineate some useful guidelines for upper bound derivations.

\subsection{Lower Bound Techniques}
\subsubsection*{Reduction to Communication Complexity}


It is often possible to reduce a finite memory inference problem to a communication complexity problem, essentially replacing the bounded memory constraint by a bounded communication constraint. The high level idea is as follows: Suppose we have an $S$-state algorithm that can solve the inference problem given $n$ samples, within the desired level of risk. Now, we can partition the samples into two parts $a$ and $b$, giving $a$ to Alice and $b$ to Bob. Alice can run the algorithm on $a$, and send the final state to Bob using $\log{S}$ bits. Bob can then start his machine at Alice's last state, and continue running it on $b$, thereby emulating the entire inference procedure. One can now try to find some function $f(a,b)$ such that solving the inference problem also allows Bob to compute $f(a,b)$ with good probability. The communication complexity of reliably computing $f(a,b)$, i.e., the minimal number of bits Alice and Bob need to exchange to that end, is therefore a lower bound on $\log{S}$, and can sometimes be evaluated directly. We note that this useful reduction originates from the streaming literature, particularly from the seminal work of Alon et al.~\cite{alon1999space} on the memory complexity of approximating the frequency moments of a finite sequence, a paper that initiated the research on the memory complexity of data streams statistics approximation.

Following the footsteps of~\cite{alon1999space}, many subsequent works applied the tools of communication complexity to derive lower bounds on different problems in the streaming model, see, e.g.,~\cite{indyk2003tight,woodruff2004optimal,woodruff2007efficient,kane2010exact,jayram2013optimal}. Applications of this reduction approach in the memory-limited statistical inference literature are few and far between. Among those are the works of Woodruff~\cite{woodruff2009average} on counting distinct elements when the samples are drawn from a particular family of distributions; Crouch et al.~\cite{crouch2016stochastic} on approximating the collision probability and other memory limited problems; Dagan and Shamir~\cite{dagan2018detecting} on detecting correlations with little memory and communication; and Dinur~\cite{dinur2020streaming} on distinguishing between sampling with or without replacement.

\subsubsection*{The Information Theoretic Approach}

Information-theoretic based lower bounds for the memory constrained inference problem typically rely on lower bounding the mutual information between $\pi$, the property we want to infer, and the memory state (or, similarly, bounding the statistical distance between the algorithm output distributions for different values of $\pi$). Capturing the tradeoff between how large this mutual information can be, and the number of available memory states, can be thought of as a variant of the information bottleneck problem~\cite{tishby2000information,goldfeld2020information}.
Though there are some examples that leverage this approach directly (see, e.g.,~\cite{diakonikolas2019communication,shahaf2020information,jaeger2019tight}), in general it is quite difficult to quantify the information contraction or find the strong data processing coefficient for many memory-limited problems. Thus, a more simple (yet significantly looser) information theoretic approach seeks to derive bounds for a communication constrained problem known as the \textit{blackboard} model: This is a broadcast model of communication
in which each sample is observed by a different player, and each of the players is allowed to communicate at most some constant number of bits to all other players. The communication constraints do not allow for arbitrary communication, but rather only one-pass protocols in which all information is communicated from left to right. 
This corresponds to a much less restrictive memory-constrained algorithm whose output at time $t$ is a function of all memory states up to time $t$ (not just the current memory state, as in our model), so the lower bounds for the blackboard problem carry over to our model. 
In the statistical learning setup, this approach was used to obtain memory-sample trade-offs by Shamir~\cite{shamir2014fundamental}, who gave lower bounds on the memory needed for a few learning and estimation problems by reducing them to a canonical 'hide and seek' problem, and was also used by Steinhardt et al.~\cite{steinhardt2016memory} to obtain lower bounds for memory constrained sparse linear regression. 

\subsubsection*{Analysis of Finite Markov Chains}

Another approach is to leverage the fact that any finite state machine driven by an i.i.d. sequence induces a homogeneous Markov chain behaviour. Thus, for time-invariant algorithms where the number of processed samples is sufficiently large w.r.t the number of states, one can expect the chain to converge to some limiting distribution, which allows us to obtain lower bounds by showing that the dependence of that limiting distribution on the property we want to infer is not too strong. We note that this approach breaks down for a time-varying algorithm, since the number of states grows with time and hence analyzing the limiting distribution becomes intractable. 

An early example of this method appears in the seminal paper of Hellman and Cover~\cite{hellman1970learning} on memory-limited binary hypothesis testing. There, the authors showed that the state likelihood ratio between the stationary distributions under the two hypotheses cannot change too quickly between states, and leveraged this to obtained a lower bound on the achievable error probability.
A more recent example is the work of Berg et al.~\cite{berg2020binary} on binary hypothesis testing with memory limited deterministic machines. There, the authors build on probabilistic pigeon-hole arguments that capture the fact that different distributions have a non-negligible probability of traversing similar Markov trajectories. In particular, they showed that for any finite-state machine, there must exist a ``bottleneck'' state that results in large
errors of both types, and is visited with non-negligible probability under both hypotheses. 

\subsection{Upper Bound Techniques}\label{sec:bin_reduce}
Loosely speaking, to construct good upper bound, we want the Markov chain induced by our finite state algorithm to have the following high-level properties:
\begin{itemize}
    \item ``Good'' limiting distribution, e.g., strongly influenced by the input distribution, so that small changes in the property we want to infer would result in large changes in the limiting distribution.
    \item Fast convergence to the limiting / stationary distribution, that is, short mixing time. Such behavior will guarantee that a small number of samples suffice for approaching the limiting distribution, which, in turn, results in a small sample complexity.
\end{itemize}

Though we are unaware of a general recipe for algorithms that satisfies both properties above, a couple of approaches that were used in the literature may be illuminating: First, when memory is large enough (say $\gtrapprox \mathsf{SC}(\delta)\cdot \log|\m{X}|$ in bits), an appealing and simple approach to deriving upper bounds is to realize the existing ``natural'' algorithm from the unconstrained setup in a memory limited scheme. On the other hand, when samples are abundant, the approach of waiting for extreme events has been useful in formulating order-wise optimal (or close to optimal) algorithms, as we expand on in later sections. 

It is worth mentioning that, in some particular problems, essentially only the ``natural'' or the rare events approach to algorithmic construction needs to be considered. For a concrete example, in the seminal work of Raz~\cite{raz2016fast}, the author addressed the problem of parity learning over $\{0,1\}^d$ with finite-memory algorithms, and showed that any algorithm for parity learning requires either a quadratic memory size in bits, or an exponential number of samples. Consequently, if one has $O(d^2)$ memory bits, parity learning can be solved in polynomial time by Gaussian elimination, which is the natural approach. Conversely, if one has an exponential number of samples, parity learning can be solved by waiting for all the singletons, consuming only $d+o(d)$ memory bits, which would be a rare events approach. 

\subsubsection*{Waiting for Rare Events}
In settings where the number of available samples is arbitrarily large, waiting has no cost, thus it makes sense to focus on capturing suitable rare events in the sample space such that, conditioned on such events, the inference task becomes easier. The memory size is then related to the rarity of the event, since we need to have sufficiently large memory in order to detect it in the first place, but describing a few rare events usually does not take much memory.  This type of approach has been originally suggested by Cover~\cite{cover1969hypothesis} for the binary hypothesis testing problem, and has been studied by others since then. In the construction of Hellman and Cover~\cite{hellman1970learning}, the finite state machine only changes a state upon observing a maximal or minimal likelihood ratio event. When it reaches an extreme state it also waits for an extreme event, but then exits the state with small probability, due to the use of artificial randomization. The deterministic binary hypothesis testing scheme of Berg er al.~\cite{berg2020binary} is based on observing a long consecutive run of $0$’s or a long consecutive run of $1$’s, which have exponentially small probabilities, but conditioned on these events, one hypothesis becomes extremely more likely than its counterpart. Specifically, both rare events schemes described above attain a risk of $2^{-\Omega(S)}$. This should be juxtaposed with using the natural statistic for this problem, the number of $1$’s in the sample, which leads to poor performance; the quadratic risk cannot decay faster than $2^{-\Omega(\sqrt{S})}$. This follows since in order to count the number of $1$’s in a stream of length $k$ we must keep a clock that counts to $k$, thus overall the number of states used is $S=O(k^2)$. As it turns out, the deterministic binary hypothesis testing machine that seeks long repetitions can be used as a building block to construct
optimal machines in other settings (see~\cite{berg2021deterministic,berg2022memory}), thus it seems that the rare event approach is fundamental for finite memory problems where the number of samples is very large.

\subsubsection*{Reduction to Binary Problems}
The approach of breaking down a complex problem into a collection of smaller simpler tasks has always been prominent in engineering and computer science. This approach turns out to be quite useful in the memory-limited context, where reducing the complexity of the problem often results in a major reduction in memory consumption. When the number of samples is not a bottleneck, one can sometimes break down the problem into a sequence of simpler, often even binary problems, using only a small number of memory states at each stage, resulting in savings to the total memory usage. The idea is, in each step, to partition the universe into two parts, and test whether we are likely to be in one or the other. It is somewhat reminiscent to noisy binary search (see, e.g.,~\cite{horstein1963sequential,nowak2008generalized,shayevitz2011optimal}), with the distinction that due to memory limitations one cannot deploy the optimal search strategy.

The earliest work adopting this approach in the memory-constrained framework seems to be that of Wagner~\cite{wagner1972estimation}, who constructed a time-varying algorithm for the estimation of the bias of a coin by decomposing the problem to a sequence of binary hypothesis testing sub-problems, leveraging the results of Cover~\cite{cover1969hypothesis}. A similar approach was recently taken by Berg et al.~\cite{berg2021deterministic}, who constructed a deterministic time-invariant bias estimation algorithm that breaks down the problem to deterministic time-invariant binary hypothesis testing sub-problems. In a recent work by the same authors~\cite{berg2022memory}, the problem of \textit{uniformity testing} (in which we test whether the underlying distribution is uniform or far from it) was reduced to a sequence of binary hypothesis tests as well, such that a success in all tests is very indicative of the uniform distribution hypothesis.  We expand on these results in the sequel.  

\section{Simple Hypothesis Testing}\label{sec:hyp}
One of the earliest appearances of a memory constraint problem in the statistical inference literature is in the hypothesis testing context. Recall that in this problem, i.i.d. samples are drawn from an unknown distribution $P \in \m{F}=\{P_0,\ldots,P_{d-1}\}$, where under hypothesis $\m{H}_i$ we have $P=P_i$, and the property we want to guess is the true underlying hypothesis,  $\pi(P_i)=i$, under the $0/1$ loss, $\ell (\pi,\hat{\pi})=\ind(\pi \neq \hat{\pi})$. For the majority of this section we will be interested in the \textit{binary hypothesis testing} problem, in which $\m{F}=\{P_0,P_1\}$. We show that even this simple version of maybe the most basic inference task is still difficult under memory limitations. In most of the section, we will be interested in the limit of the problem when $n\rightarrow \infty$, thus following Eq.~\eqref{eq:infinity_risk}, we define the asymptotic probability of error of an $S$-state algorithm, when the true hypothesis is $\mathcal{H}_i$, to be
\begin{align}
   \Pe (P_i,\hat{\pi})=\limsup_{n\rightarrow\infty}\E[\ind(\hat{\pi}(X^n)\neq \pi(P_i) )]=\limsup_{n\rightarrow\infty}\Pr(\hat{\pi}(X^n)\neq i). 
\end{align}
Our main interest is thus to characterize the \textit{asymptotic minimax error probability}
\begin{align}
    \mathsf{P^*_e}(S)\triangleq \inf_{\hat{\pi}\in \fsm(S)}\sup_{P\in \{P_0,P_1\}}\Pe (P,\hat{\pi}).
\end{align}
Interest in the limited memory hypothesis testing problem seems to have been sparked by the work of Robbins~\cite{robbins1956sequential} on the Two-Armed Bandit problem. In this setup, a player is given two coins, with Bernoulli parameters $p_1$ and $p_2$ respectively, whose values are unknown. The player is required to maximize the long-run proportions of ``heads'' obtained, by successively choosing which coin to flip at any moment. Robbins proposed an algorithm that works with a limited memory of $S$ states and attains the optimal performance only as $S$ approaches infinity. Robbin's algorithm was later improved by Isbell~\cite{isbell1959problem}, Smith and Pike~\cite{smith1965robbins}, and Samuels~\cite{samuels1968randomized}, where the latter improvement arises from a randomized algorithm.

Departing from the restriction to time-invariant algorithms imposed in the works mentioned above, Cover~\cite{cover1968note} discovered a time-varying finite memory algorithm that achieves the optimal proportion $\max\{p_1,p_2\}$, asymptotically in the number of tosses, with $S=2$. Essentially, Cover suggested to modify the problem by allowing to keep a clock, such that at each stage we know how many tosses we have made so far. Cover's algorithm interleaves test blocks and trial blocks, where a test block is a sequence of tosses that test the hypothesis $p_1>p_2$ vs. $p_2>p_1$, and a trial block is a sequence of exclusive tosses of the ``favorite'' coin resulting from the preceding test block. The choice of the sequence of test-trial block sizes then assures that the maximum is attained. The intuitions and tools developed for the Two-Armed Bandit problem were used by Cover in a subsequent paper addressing the binary hypothesis problem~\cite{cover1969hypothesis}, in which he described a time-varying finite memory machine that solves the binary hypothesis testing problem with $S=4$ states (this was later improved to $S=3$ by Koplowitz~\cite{koplowitz1975necessary}, who more generally showed that for $M$-ary hypothesis testing with time-varying finite memory, $M + 1$ states are necessary and sufficient to resolve the correct hypothesis with an error probability that approaches zero asymptotically). Specifically, Cover describes an algorithm with four memory states that, given samples from a $\Bern(p)$ distribution, can decide between the hypothesis $\m{H}_0:p=p_0$ and $\m{H}_1:p=p_1>p_0$ with error probability that converges to zero almost surely.\footnote{We note that the result in the paper is stronger, solving a composite hypothesis testing problem of the form $\m{H}_0:p>p_0$ vs.
$\m{H}_1:p<p_0$.}

The algorithm works as follows. We keep two binary random variables $(T,Q)$, where $T$ keeps the current guess of the correct hypothesis, and $Q$ keeps the result of the current run test, to be immediately defined. The sequence of Bernoulli samples is then divided into blocks $S_1,R_1,S_2,R_2,\ldots$ with corresponding lengths $s_1,r_1,s_2,r_2,\ldots$, where in each $S_i$ block we test for a run of $s_i$ consecutive zeros, in which case the $S_i$ block is considered a success, and in each $R_i$ block we test for a run of $r_i$ consecutive ones, in which case the $R_i$ block is considered a success. The value of $Q$ tracks after the current run, by being set to $1$ as long as the run continues and being set to $0$ for the rest of the block if it breaks. Only if the test succeeded, the value of the current hypothesis guess $T$ is updated to that of the tested hypothesis at the end of the test block, otherwise it retains its previous value. The probability of success in an $S_i$ block is thus $p^{s_i}$, and the probability of success in an $R_i$ block is $(1-p)^{r_i}$. By setting $s_i\approx \log_{p_0}(1/i)$ and, similarly, $r_i\approx \log_{1-p_0}(1/i)$, we have that $p>p_0$ implies that both $\sum p^{s_i}=\infty$ and $\sum (1-p)^{r_i}<\infty$, so from Borel–Cantelli lemma we have that $T_n\rightarrow 1$ with probability $1$, and, similarly, for $p<p_0$ we have that $T_n\rightarrow 0$ with probability $1$. 

In the paper itself, Cover states that \say{\textit{However, the dependence of $f_n$ on $n$ requires external specification of $n$ if the algorithm is to be considered to have truly finite memory.}} This point is important; without an external input, a time-varying algorithm is required to keep a clock, which results in unbounded memory under the asymptotic framework. Thus, it would seem natural that, to truly understand the memory limits of the hypothesis testing problem, we must restrict ourselves to time-invariant algorithms, thus taking into account any use of clock in the memory budget. 
\subsection{Randomized Binary Hypothesis testing}
Hellman and Cover~\cite{hellman1970learning} addressed the problem of binary hypothesis testing within the class of randomized time-invariant finite-state machines, and found an \textit{exact}
characterization of the smallest attainable error probability for this problem.\footnote{In the paper, the authors discuss the Bayesian setting, in which each hypothesis is assigned a prior probability. For consistency, we present the result for the minimax setting, which coincides with their result when the prior is equiprobable.} Specifically, define the maximum and minimum likelihood ratios as
\begin{align}
   \overline{L}=\max _{A\in \m{X}}\frac{P_0(A)}{P_1(A)}\hspace{1mm}; \hspace{2mm} \underline{L}=\min _{A\in \m{X}}\frac{P_0(A)}{P_1(A)},
\end{align}
and define $\gamma=\frac{\overline{L}}{\underline{L}}$. It holds that
\begin{align}
    \mathsf{P^*_e}(S)=\frac{1}{\gamma^{\frac{1}{2}(S-1)}+1}.~\label{eq:opt_err_prob}
\end{align}

Hellman and Cover showed that the error probability of any algorithm cannot be smaller than~\eqref{eq:opt_err_prob}, and that, for any $\veps>0$, there exists an $S$-states randomized algorithm that achieves error probability of $\mathsf{P^*_e}(S)+\veps$. As the optimal error probability decreases exponentially with $S$, we can define the asymptotics of the error exponent with regards to
$S$ as
\begin{align}
   \lim_{S \rightarrow \infty}-\frac{1}{S}\log \Pe^{*}(S) =\frac{1}{2}\cdot \log \gamma.
\end{align}
We call the above the \textit{randomized error exponent} of the binary hypothesis testing problem with finite memory. 
\subsubsection*{Hellman and Cover~\cite{hellman1970learning} - Lower bound} 

For the lower bound, first assume that the Markov process induced by the finite-state algorithm is irreducible. There is no loss of generality in that assumption, as it is proved that the error probability of any $(S+1)$-state algorithm that induces a reducible Markov process is lower bounded by $\mathsf{P^*_e}(S)$. For such processes, there is a unique stationary distribution induced by the input distribution. Denote by $\mu_i^{0}$ (resp. $\mu_i^{1}$) the stationary probability of state $i$ in the chain, under hypothesis $\mathcal{H}_0$ (resp. $\mathcal{H}_1$), and define the \textit{state likelihood ratios} (SLR) $\lambda_i \triangleq \mu_i^0 / \mu_i^1$. A low error probability intuitively implies that $\lambda_i \ll 1$ or $\lambda_i \gg 1$ for most $i \in [S]$, so to get a bound that holds for any machine we first need to show that the state likelihood ratio cannot change too much. Specifically, assuming w.l.o.g. that the SLR's are non-decreasing, it is shown that 
\begin{align}
   1\leq \frac{\lambda_{i+1}}{\lambda_i}\leq \gamma .\label{eq:ratio}
\end{align}
The lower bound follows from the assumption and the upper bound follows from a straightforward analysis of Markov chain stationary distributions, using the fact that the most distinguishing observation is the one the maximizes the likelihood ratio, thus our state likelihood ratio cannot grow by more than $\gamma$. Let $\alpha$ and $\beta$ be the error probabilities under hypothesis $\m{H}_0$ and $\m{H}_1$, respectively. The worst case error probability is thus $\Pe(S)=\max \{\alpha,\beta\}$. If $\lambda_{\min}$ is the minimal state likelihood ratio, then for all $i\in [S]$, Eq.~\eqref{eq:ratio} implies that
\begin{align}
   \lambda_{\min} \leq \frac{\mu_i^0}{\mu_i^1} \leq \lambda_{\min}\cdot\gamma^{S-1}.
\end{align}
Let $\m{S}_0$ denote the decision region for hypothesis $\m{H}_0$, i.e., the set of states in $[S]$ that are mapped to the estimate $\hat{\pi}=0$. Similarly, let $\m{S}_1$ denote the decision region for hypothesis $\m{H}_1$. Now, writing the error probability for each of the hypotheses, we get the bounds
\begin{align*}
    \alpha&=\sum_{i\in \m{S}_1}\mu_i^0\geq \lambda_{\min}\sum_{i\in \m{S}_1}\mu_i^1 = \lambda_{\min}(1-\beta)\\ \beta&=\sum_{i\in \m{S}_0}\mu_i^1\geq \frac{1}{\lambda_{\min}\cdot\gamma^{S-1}}\sum_{i\in \m{S}_1}\mu_i^0 = \frac{1}{\lambda_{\min}\cdot\gamma^{S-1}}(1-\alpha).
\end{align*}
Multiplying both equations we get $\alpha\beta\geq\gamma^{-(S-1)}(1-\alpha)(1-\beta)$, a constraint on the achievable error probabilities under both hypotheses, and minimizing $\Pe(S)$ under the constraint yields the result.

\subsubsection*{Hellman and Cover~\cite{hellman1970learning} - Upper bound}
The upper bound construction can be understood by the following intuition: Assume that the hypothesis testing problem is dealing with the Bernoulli symmetric case, where the samples arrive either from a $\Bern(p)$ or a $\Bern(1-p)$ distribution. Consider a \textit{saturable counter} machine, which is a machine with transition rule that moves up upon seeing $1$, moves down upon seeing $0$, and either stays in place or moves back when it arrives at the extreme states (called saturated states). Furthermore, let the decision rule map each state to the hypothesis with the larger stationary distribution. A simple math exercise shows that the error probability we obtain decreases exponentially in the number of states, but at a rate that does not attains the lower bound described above. 

The ingenious modification of Hellman and Cover is to move more probability mass to the saturated states, where an error is less likely. To achieve this, the algorithm only stays back from saturation when, in addition to the input sample pointing the other way, an external $\Bern(\delta)$ random coin falls on head. The parameter $\delta$ is typically very small, and its value depends on $p$. Extending it to the general binary hypothesis testing problem (not necessarily Bernoulli symmetric), the machine now moves up on high likelihood-ratio events, moves down on low likelihood-ratio events, and stays put otherwise. In the extreme state corresponding to hypothesis $\m{H}_0$, the process moves back upon seeing a low likelihood-ratio event with probability $\delta$, chosen according to $\gamma$. In the extreme state corresponding to hypothesis $\m{H}_1$, the process moves back upon seeing a high likelihood-ratio event with probability $k\cdot\delta$, where $k$ is chosen to account for the asymmetry between the hypotheses (see Figure~\ref{fig:helmancover}). This defines a class of $\veps$-optimal finite state machines, i.e., for any $\veps>0$, there exists an machine in this class that achieves probability of error at most $\Pe^{*}(S)+\veps$. Thus, the optimal value can be approached arbitrarily closely.
\begin{center}
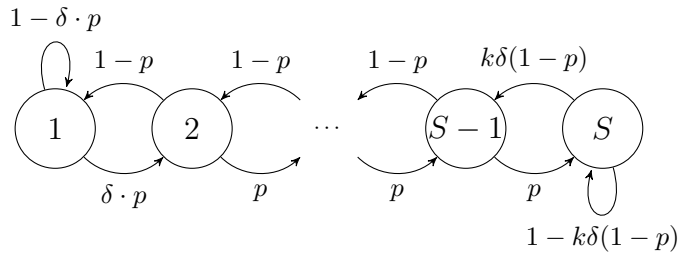
\begin{figure}[H]
\centering
\setlength\belowcaptionskip{-1.4\baselineskip}
\begin{tikzpicture}
  \tikzset{
    >=stealth',
    node distance=1.8cm,
    state/.style={font=\scriptsize,circle, align=center,draw,minimum size=30pt},
    dots/.style={state,draw=none}, edge/.style={->},
  }
  \node [state ,label=center:$1$] (1) {};
  \node [state ,label=center:$2$] (2)   [right of = 1] {};
  \node [dots]  (dots)  [right of = 2] {$\cdots$};
  \node [state,label=center:$S-1$] (S-1) [right of = dots]  {};
  \node [state,label=center:$S$] (S) [right of = S-1]  {};
  \path [->,draw,thin,font=\footnotesize]  (1) edge[loop above] node[above] {$1-\delta\cdot p$} (1);
  \path [->,draw,thin,font=\footnotesize]  (1) edge[bend right=45] node[below] {$\delta \cdot p$} (2);
  \path [->,draw,thin,font=\footnotesize]  (2) edge[bend right=45] node[above] {$ 1-p$} (1);
  \path [->,draw,thin,font=\footnotesize]  (2) edge[bend right=45] node[below] {$ p$} (dots);
  \path [->,draw,thin,font=\footnotesize]  (dots) edge[bend right=45] node[above] {$1-p$} (2);
  \path [->,draw,thin,font=\footnotesize]  (dots) edge[bend right=45] node[below] {$p$} (S-1);
  \path [->,draw,thin,font=\footnotesize]  (S-1) edge[bend right=45] node[above] {$1-p$} (dots);
  \path [->,draw,thin,font=\footnotesize]  (S-1) edge[bend right=45] node[below] {$ p$} (S);
  \path [->,draw,thin,font=\footnotesize]  (S) edge[bend right=45] node[above] {$k\delta (1-p)$} (S-1);
 \path [->,draw,thin,font=\footnotesize]  (S) edge[loop below] node[below] {$1-k\delta (1-p)$} (S);
\end{tikzpicture}
\caption{Randomized binary hypothesis testing machine of~\cite{hellman1970learning} for $\Bern(p)$ input}\label{fig:helmancover}
\end{figure}
\end{center}

\subsection{Deterministic Binary Hypothesis testing}~\label{sec:det_hyp}
As stated previously, the binary hypothesis testing algorithm that achieves the Hellman-Cover lower bound is randomized. In the following, we restrict our discussion to deterministic algorithms. For simplicity, assume we observe either the $\Bern(p)$ distribution, under hypothesis $\mathcal{H}_0$, or the $\Bern(q)$ distribution, under hypothesis $\mathcal{H}_1$, for $0<q<p<1$. Since the optimal error probability decreases exponentially in $S$, we can define the error exponent as
\begin{align}
  \mathsf{E}(p,q)=-\lim_{S \rightarrow \infty} \frac{1}{S}\log \Pstar(S) =\frac{1}{2}\cdot \log \gamma = \frac{1}{2}\log \frac{p(1-q)}{q(1-p)}. 
\end{align}
Now letting $\Pstar_{\text{,det}}(S)$ be the optimal error probability that can be attained using deterministic algorithms, we want to similarly characterise the \textit{deterministic error exponent},
\begin{align}
    \mathsf{E}_{\mathsf{det}}(p,q)=-\lim_{S \rightarrow \infty} \frac{1}{S}\log \Pstar_{\text{,det}}(S).
\end{align}

As deterministic machines are a special case of randomized machines, we clearly have $\mathsf{E}_{\mathsf{det}}(p,q)\leq \mathsf{E}(p,q)$.
To demonstrate the important role randomization plays in approaching the optimal performance, Hellman and Cover showed in~\cite{hellman1971memory} that for any memory size $S<\infty$ and $\delta>0$ there exists problems such that any $S$-state deterministic machine has probability of error $\Pe\geq \frac{1}{2}-\delta$, while the randomized machine from~\cite{hellman1970learning} has $\Pe\leq \delta$. In~\cite{hellman1972effects} (see also~\cite{hellman1973review}) it was shown that $\mathsf{E}_{\mathsf{det}}(p,q)$ is positive for all $p\neq q$, and for the symmetric setting, where $p=1-q$, Shubert et al.~\cite{shubert1973testing} also derived a lower bound on $\mathsf{E}_{\mathsf{det}}(p,q)$. This implies that whenever $\gamma<\infty$, i.e., for any $0<p,q<1$, there exists some integer  $1\leq C=C(p,q)<\infty$ such that $\Pstar_{\text{,det}}(S\cdot C)\leq \Pstar(S)$, \emph{for all $S$}. 

The problem has been largely abandoned at that point, until fairly recently, when Berg et al.~\cite{berg2020binary} analyzed the error probability of optimal deterministic machines, and gave upper and lower bounds on $\mathsf{E}_{\mathsf{det}}(p,q)$ which coincide for some extreme values of the hypotheses. Specifically, they showed that the optimal error probability of a deterministic machine has
\begin{align}
   r(p,q)\leq\mathsf{E}_{\mathsf{det}}(p,q)\leq d(p,q),
\end{align}
where
\begin{align}
   r(p,q)&=\frac{\log p \log (1-q)-\log q \log (1-p)}{\log p(1-p)+\log q(1-q)},\\ d(p,q)&=\frac{\log(\min \{p,1-p\})\cdot \log(\min \{q,1-q\})}{\log(\min \{p,1-p\})+\log(\min \{q,1-q\})}. 
\end{align}

The lower bound on the error exponent is comprised of an algorithm that waits for events that very sharply distinguish between the hypotheses, specifically, a long consecutive
run of either zeros or ones, where the run length is chosen to optimize the error probability (see Figure~\ref{fig:RUNS}). This $N$-state algorithm for testing between the hypotheses $\m{H}_0=p$ and $\m{H}_1=q$ is referred to as $\isit(N,p,q)$, and it achieves error probability $\delta$ whenever $p=q+\veps$ if the number of states is $N= O(1 / \veps \cdot \log (1/\delta ))$. 
\begin{center}
\begin{figure}[H]
\centering
\setlength\belowcaptionskip{-1.4\baselineskip}
\begin{tikzpicture}
  \tikzset{
    >=stealth',
    node distance=0.92cm,
    state/.style={font=\scriptsize,circle, align=center,draw,minimum size=23pt},
    dots/.style={state,draw=none}, edge/.style={->},
  }
  \node [state ,label=center:$1$] (S0)  {} ;
  \node [state] (S0-1)   [right of = S0]   {};
  \node [dots] (dots1)   [right of = S0-1]   {$\cdots$};
  \node [state] (1l) [right of = dots1]  {};
  \node [state ,label=center:$s$] (0)   [right of = 1l] {};
  \node [state] (1r) [right of = 0]   {};
  \node [dots]  (dots2)  [right of = 1r] {$\cdots$};
  \node [state] (S1-1) [right of = dots2]  {};
  \node [state ,label=center:$N$] (S1) [right of = S1-1]  {};
  \path [->,draw,thin,font=\footnotesize]  (S0-1) edge[bend left=45] node[below ] {$1-p$} (S0);
  \path [->,draw,thin,font=\footnotesize]  (0) edge[bend left=45] node[below] {$1-p$} (1l);
  \path [->,draw,thin,font=\footnotesize]  (1r) edge[bend left=45] node[below right] {$1-p$} (1l);
  \path [->,draw,thin,font=\footnotesize]  (S1-1) edge[bend left=45] node[below right] {$1-p$} (1l);
  \path [->,draw,thin,font=\footnotesize]  (1l) edge[bend left=45] node[below ] {$1-p$} (dots1);
  
  \path [->,draw,thin,font=\footnotesize]  (S0-1) edge[bend left=45] node[above left] {$p$} (1r);
  \path [->,draw,thin,font=\footnotesize]  (1l) edge[bend left=45] node[above left] {$p$} (1r);
  \path [->,draw,thin,font=\footnotesize]  (0) edge[bend left=45] node[above ] {$p$} (1r);
  \path [->,draw,thin,font=\footnotesize]  (1r) edge[bend left=45] node[above] {$p$} (dots2);
  \path [->,draw,thin,font=\footnotesize]  (S1-1) edge[bend left=45] node[above ] {$p$} (S1);
 \path [->,draw,thin,font=\footnotesize]  (S0)  edge[loop left]  node[above ]{$1$} (S0);
 \path [->,draw,thin,font=\footnotesize]  (S1)  edge[loop right] node[above ]{$1$} (S1);
\end{tikzpicture}
\caption{$\isit(N,p,q)$ - Deterministic Binary Hypothesis Testing Machine}		\label{fig:RUNS}
\end{figure}
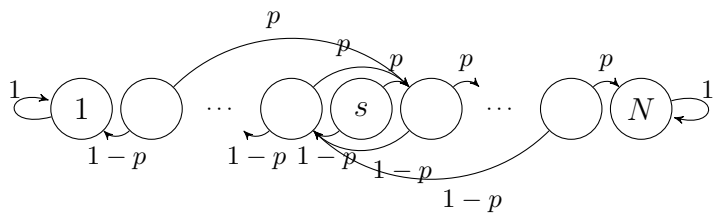
\end{center}
The upper bound is based on the insight that, while in randomized schemes the transition probabilities between states can be as small as desired, in deterministic machines the transition
probabilities can only be as small as $\min\{p,1-p\}$ or $\min\{q,1-q\}$. More specifically, in order to derive a lower bound on $\Pstar_{\text{,det}}(S)$ for ergodic machines, it is first shown that the error probability of any algorithm is lower bounded by $\max_{i\in[S]}\min \{\mu_i^p,\mu_i^q\}$, where $\mu_i^p$ is the stationary distribution of state $i$ when the samples are drawn from the $\Bern(p)$ distribution. It is then left to show that there exits a state $i\in [S]$ for which both $\mu_i^p$ and $\mu_i^q$ are large. The proof follows from the observation that, if state $i$ is connected to state $j$, we must have $\mu_j^p\geq \mu_i^p\cdot \min \{p,1-p\}$. Using the argument recursively, and applying it to $\mu_i^q$ as well, gives the bound. The authors then consider the two absorbing states case, one for each competing hypothesis. It is shown that in any finite state machine, there must exist some state that is highly likely under both hypotheses and is close to both absorbing states, and this forces the error probability to be large as well. The general non ergodic case is proven by a superposition of the ergodic and two absorbing states cases.

Though in general not tight, the lower bound demonstrates the gap between the error exponent for deterministic machines and that of randomized ones, which was derived in~\cite{hellman1970learning}. In particular, for certain values of $(p,q)$, it is shown that the restriction to deterministic machines may arbitrarily degrade the error exponent. Specifically, for any fixed $q<\frac{1}{2}$, we have $\lim_{p \rightarrow 1}\mathsf{E}_{\mathsf{det}}(p,q)=-\log q$ while  $\lim_{p \rightarrow 1}\mathsf{E}(p,q)=\infty$, and, similarly, for any fixed $p>\frac{1}{2}$, we have $\lim_{q \rightarrow 0}\mathsf{E}_{\mathsf{det}}(p,q)=-\log (1-p)$ while  $\lim_{q \rightarrow 0}\mathsf{E}(p,q)=\infty$.

\subsection{Parity Learning}
Here the family of distributions $\m{F}$ is characterized by some $\theta\in \{0,1\}^d$ and is given as a pair of observations $(X,Y)$, where $X\sim \Bern^{\otimes d}(1/2)$ and $Y=X^T \theta$, where $ X^T \theta$ denotes inner product modulo 2. In particular, we observe samples $(X_1, Y_1), (X_2, Y_2)\ldots$, where each $X_t$ is uniformly distributed over
$\{0,1\}^d$ and for every $t$, $Y_t = X_t^T \theta$ and we want to estimate the vector $\theta$ under the $0-1$ loss.

Following a conjecture of Steinhardt et al.~\cite{steinhardt2016memory}, Raz proved in a seminal paper~\cite{raz2016fast} that the problem of parity learning provides the following separation: Any algorithm for parity learning that uses less than $d^2/25$ bits of memory requires at least $2^{d/25}$ samples. To appreciate this result, assume that $\delta$ is some small constant. First note that for the parity learning problem we have $\sc(\delta)\asymp d$ and $\mc (\delta)\asymp d$ in bits. The sample-complexity follows since each observation gives at most $1$ bit of information on $\theta$, and $O(d)$ time can be achieved by using the natural approach of simply saving all the equations as if there are no memory limitations using $O(d^2)$ memory bits, and then performing Gaussian elimination. The memory-complexity follows since we need at least $d$ bits to represent $\theta$, and we can achieve $O(d)$ by the rare events approach of waiting for all the singletons, which takes an exponential number of samples in $d$. The results of~\cite{raz2016fast} can be translated as a sharp decay of $n^*(S,\delta)$ on the sample-memory curve, around $S\asymp d^2$ bits, from exponential in $d$ to linear in $d$. 

Building on this result, Kol et al.~\cite{kol2017time} considered the problem of learning \textit{sparse} parities. They showed that the sparse parity learning problem, over the set of all vectors of Hamming
weight exactly $\ell$, requires either a memory of size super-linear in $d$ or a number of samples super-polynomial in $d$, as long as $\ell \geq \omega(\log d / \log\log d)$.
The work of Raz~\cite{raz2017time} expanded these previous results by showing that the same sharp drop-off of $n^*(S,\delta)$ in the sample memory curve is not only inherent to the parity learning problem, but is exhibited by a large class of problems. Specifically,  let $\Theta,\m{X}$ be two finite sets, and let $M : \Theta \times \m{X} \rightarrow\{-1, 1\}$ be a matrix. The learning problem is the following: An unknown element $\theta \in \Theta$ is chosen uniformly at random. A learner then tries to learn $\theta$ from a stream of samples, $(X_1, Y_1),(X_2, Y_2),\ldots$, where for every $i, X_i \in \m{X}$ is chosen uniformly at random and $Y_i = M(X_i, \theta)$.
Let $\sigma_M$ be the largest singular value of $M$ and note that $\sigma_M\leq |\m{X}|^{\frac{1}{2}}\cdot |\Theta|^{\frac{1}{2}}$, since the spectral norm is upper bounded by the Frobenius norm. The paper shows that if $\sigma_M\leq |\m{X}|^{\frac{1}{2}}\cdot |\Theta|^{\frac{1}{2}-\veps}$, then any learning algorithm for the corresponding learning problem requires either a memory of at least $\Omega((\veps d)^2)$ bits or at least $2^{\Omega(\veps d)}$ samples, where $d=\log |\Theta|$.

In a recent work, Garg et al.~\cite{garg2021memory} leveraged the tools developed in~\cite{raz2016fast} to give memory-sample lower bounds for learning parity with noise. In particular, they showed that when a stream of random linear equations over $\mathbb{F}_2$ 
is of the form $(X_1,\theta^TX_1\oplus Z_1),\ldots,(X_n,\theta^TX_n\oplus Z_n)$ where $\{Z_i\}_{i=1}^n$ are distributed i.i.d $\Bern(\veps)$ for some $0<\veps<1/2$, then the drop-off of $n^*(S,\delta)$ in the sample-memory curve occurs around $S\asymp d^2/\veps$ bits. 


\subsubsection*{Lower Bound of Raz~\cite{raz2016fast}}

The works described above rely heavily on analyzing the evolution of the posterior on $\theta$ along the progress of a \textit{branching program}. This technique was introduced in~\cite{raz2016fast}, and we will attempt to elucidate it now. A branching program of length
$n$ and width $S$ is just a graphic representation of a time varying FSM with $S$ states, operating on $n$ samples. It is represented as a directed graph with vertices arranged in $n+1$ layers
containing at most $S$ vertices each. Each layer represents a time step and each vertex represents a memory state of the learner. In the first layer, that we think of as layer $0$, there is only one vertex, called the start vertex. A vertex of outdegree $0$ is called a leaf. All vertices in the last
layer are leaves (but there may be additional leaves). Every non-leaf vertex in the program has $2^{d+1}$ outgoing edges, one for each pair of values $(X,Y)=(x, y)\in \{0, 1\}^d \times\{0, 1\}$, with exactly one edge labeled by each such $(x, y)$, and all these edges going into vertices in the next layer. Each leaf $v$ in the program is labeled
by an element $\hat{\theta}(v)\in \{0,1\}^d$, that we think of as the output of the program on that leaf. The samples $(X_1, Y_1),\ldots,(X_n, Y_n) \in \{0,1\}^d\times \{0,1\}$ that are given as input define a computation-path in the branching program, by starting from the start vertex and following at step $t$ the edge labeled by $(x_t, y_t)$, until reaching a leaf. The program outputs the label $\hat{\theta}(v)$ of the leaf $v$ reached by the computation-path. The success probability of the program is the probability that $\hat{\theta} = \theta$, where $\hat{\theta}$ is the label of the vertex reached by the path, and the probability is over $\theta, X_1,\ldots, X_n, Y_1,\ldots, Y_n$.

We give a sketch of the proof of~\cite{raz2016fast}, that shows that any branching program with width $S\leq 2^{d^2/25}$ and length $n\leq 2^{d/25}$ outputs the correct $\theta$ with exponentially small probability. We assume  that each vertex $v$ remembers a set of linear equations $E_v$ such that if the computation path reaches $v$, then all equations in $E_v$ are satisfied. This assumption restricts the branching program to be in a sub-class called  $\textit{affine branching programs}$, but the author shows that any branching program for parity learning can be simulated by an affine branching program for parity learning (we skip the proof of this reduction).
The output is then a random element $\hat{\theta}\in \{0,1\}^d$ that satisfies $E_v$. The proof shows that the probability that the computation path reaches any $v$ such that $\dim (E_v)\geq k$, for $k=\frac{1}{5}d$, is exponentially small. Why is that important? If we want to correctly estimate $\theta$ with high probability, we must arrive at a $v^*$ with $\dim (E_{v^*})\approx d$. For any edge $(i,j)$, we have $\dim (E_j)\leq \dim (E_i)+1$ for all $i$, as we see only one new equation at each time step. Thus, in order to arrive at a $v^*$ with $\dim (E_{v^*})\approx d$, we must pass through some $v$ with $\dim (E_v)\geq k$.

Assume that $\dim (E_v)\geq k$ and let $Z_0,\ldots, Z_m$ be the vertices on the computation path and $r_i=\dim (E_v \cap E_{Z_i})$. 
Note that $\dim (E_{Z_{i+1}})\leq \dim (E_{Z_i})+1$ implies $r_{i+1}\leq r_i+1$ for all $i$. Thus, if we reach a $v$ with $\dim (E_v)\geq k$, then $r_{i+1}>r_i$ occurs at least $k$ times, so it is enough to bound $\Pr(r_{i+1}>r_i)$. For this event to occur, the new sample $(x_{i+1},y_{i+1})$ must lay in the space spanned by $E_v$, hence it can be shown that $\Pr(r_{i+1}>r_i)\leq 2^{-\Omega(d)}$. Then, by using the union bound over $S^k$ vertices, the probability to reach any $v$ with $\dim (E_v)\geq k$ is less than $S^k\cdot (2^{-\Omega(d)})^k= 2^{-\Omega(d^2)}$.

\subsection{Additional Results} 
The finite-memory simple hypothesis testing problem had an impact on other problems in the memory constrained literature. In~\cite{cover1970two}, Hellman and Cover solved the two armed bandit with
time-invariant finite memory problem by using a similar construction and proof elements from~\cite{hellman1970learning}. Horos and Hellman~\cite{horos1972confidence} allowed a confidence
to be associated with each decision, with errors being weighted according to the
confidence with which they are made. They found the
$\veps$-optimal class of rules to be deterministic. Flower and Hellman~\cite{flower1972hypothesis} examined the finite sample problem for the
Bernoulli case. They found that randomization was needed on all transitions toward the center state, i.e., on transitions from states of low uncertainty to states with
higher uncertainty. 
Continuing the exploration of the finite-sample hypothesis testing with finite memory problem, Cover et al.~\cite{cover1976optimal} established the existence of an optimal rule, found its structure for time-varying algorithms, and characterised the optimal error probability of the $2$-state finite-sample problem that distinguishes between two Gaussian distributions with a different mean. 

\section{Parameter Estimation}\label{sec:est}
Recall that in the parameter estimation setting, i.i.d. samples are drawn from some unknown distribution $P$ belonging to a given parametric family $\m{F}=\{P_{\theta}\}_{\theta \in \m{Y}}$, where typically $\m{Y} \subseteq \mathbb{R}^d$ for some $d$. The goal is to estimate the parameter $\theta$ itself under a prescribed loss function, hence in our notation we are interested in the property $\pi(P_{\theta})=\theta$. A common choice of loss function in this setting is the squared error loss $\ell(\hat{\theta},\theta)=\|\hat{\theta}-\theta\|^2$. Here also we will be interested in the limit of the problem when $n\rightarrow \infty$, thus following~\eqref{eq:infinity_risk} we define the asymptotic quadratic risk attained by this estimator when the true value of the parameter is $\theta$, to be 
\begin{align}
R(\theta,\hat{\pi}) = \limsup_{n\rightarrow\infty}  \E \left(\hat{\pi}(X^n)-\theta\right)^2.\label{eq:MSE} 
\end{align}
The main interest is thus to characterize the minimax risk $R^*(S)$ of the problem, which is the smallest asymptotic quadratic risk that can be uniformly guaranteed for all $\theta\in\m{Y}$ using an FSM with $S$ states.

The parameter estimation with finite-memory problem was first addressed by Roberts and Tooley~\cite{roberts1970estimation}, who tackled the problem of estimation under quadratic risk for a random variable with additive noise using time-varying machines. They worked under the restriction that larger observations cause transitions to higher numbered states. In some
problems (for example, the Cauchy distribution), very large
observations yield very little information, and such a rule seems to be
distinctly suboptimal. However, for Gaussian statistics, large observations
are very informative, in which case their form does make sense. Koplowitz and Roberts~\cite{koplowitz1973sequential} later demonstrated necessary and sufficient conditions for the
optimal state transition function. 

Wagner~\cite{wagner1972estimation} used rules similar to Cover's~\cite{cover1969hypothesis} to estimate the mean of a distribution. For Bernoulli observations, Wagner's scheme is very close to optimal, since its maximum absolute error is at most $\veps$, with $S=O(1/\veps)$ memory states. He achieved this result by partitioning the observations into blocks $\{S_k^j,R_k^j\}_{k=1}^{\infty}$ sequentially testing whether $p>p_j$ or $p<p_j$, for $1\leq j \leq r$, and $0=p_0<p_1<\ldots<p_r=1,r=O(1/\veps)$. By keeping the index $j$ of the favourite hypothesis and choosing the block lengths to follow the Borrel-Cantelli convergence conditions, the correct block is identified with probability $1$. This is an example of the methodology mentioned in Section~\ref{sec:bin_reduce}, of breaking down a complex problem into multiple simple binary decision problems; we will see this again in the sequel. 

Hellman~\cite{hellman1974finite} examined the problem of estimating the mean $\mu$ of a Gaussian Distribution $\m{N}(\mu,\sigma^2)$ under the mean squared error loss, where $\mu$ is drawn from a known prior distribution $p(\mu)$. To obtain a lower bound, consider the simpler quantization problem, where we pick the best  $S$-level scalar quantizer $Q(\mu)$ for $\mu$, minimizing the mean squared error $\E (Q(\mu)-\mu)^2$. Clearly, this provides a lower bound for the mean squared error attained by any $S$-state estimation algorithm. 
One might assume that this quantization lower bound is not tight, however, quite surprisingly, Hellman showed that the quantization lower bound can be approached arbitrarily closely by the following (deterministic) construction: Let $\mu_1^*<\mu_2^*<\ldots<\mu_S^*$ be the optimal quantization points of an $S$-level quantizer. Consider the machine which transits from state $i$ to state $i + 1$
whenever $X_n - (\mu_i^*+\mu_{i+1}^*)/2 \geq M$, and from state $i$ to state
$i - 1$ when $X_n - (\mu_{i-1}^*+\mu_i^*)/2 \leq -M$, and which estimates $\mu$ to be $\mu_i^*$ when it is in state $i$. Thus, the ratio between transition probabilities toward states with better $\mu$ estimates and transition probabilities toward states with worse $\mu$ estimates behaves exponentially with $M$, so that, as $M \rightarrow \infty$, the stationary distribution approaches a singleton on $\mu^*$, the closest value to $\mu$. Thus, in the Gaussian case, the mean estimation problem can be reduced to a quantization problem, that is, for any prior $p(\mu)$ on $\mu$ the minimum quadratic risk attained by an $S$-state estimator is equal to the quadratic risk attained by the optimal $S$-level quantizer.

The main reason we were able to attain the quantization lower bound in this case was that under the Gaussian distribution we sometimes, even though rarely, observe a sample that contains an arbitrarily large amount of information about the parameter. Thus, if we waited for such samples and only update our memory state upon observing them, we would converge to a state with the best estimator. 
Unlike the Gaussian distribution, the Bernoulli distribution has only two outcomes, $1$ and $0$, which contain very little information about the Bernoulli parameter. In this distribution, rare events can only occur w.r.t a sequence of samples, thus tracking these events incurs an additional memory cost. Indeed, we now turn our attention to the problem of estimating an unknown Bernoulli parameter, and show that in this case the quantization lower bound of $\Omega(1/S^2)$ \emph{cannot} be achieved.

Samaniego~\cite{samaniego1973estimating} initiated the work on the problem of estimating the parameter of a coin using time-invariant finite-memory machines. He considered a Bayesian setup in which a prior is placed on the parameter distribution, and the loss function is a generalized quadratic loss function of the form $(\hat{\theta}-\theta)^2\theta^{\beta-1}(1-\theta)^{\gamma-1}$, for some $\beta,\gamma\geq 0$. Furthermore, he restricted to machines that can only make transitions between adjacent states, which he refers to as \textit{tridiagonal} algorithms. Within the class of all
tridiagonal algorithms, a particular rule was shown to be minimax as well as locally Bayes w.r.t the uniform prior for $S\leq 30$. 
Leighton and Rivest~\cite{leighton1986estimating} followed up the work on estimating the parameter of a coin, and found an exact characterization of the smallest attainable mean squared error for this problem. Specifically, they proved that
\begin{align}
    R^*(S) \asymp\frac{1}{S},
\end{align}
or, equivalently, using the formulation of~\eqref{eq:mc}, they show that $\mathsf{MC}(\delta)  \asymp 1/\delta$.

\subsubsection*{Leighton and Rivest~\cite{leighton1986estimating} - Upper bound}
Samaniego's construction from~\cite{samaniego1973estimating} is analyzed and the mean squared error is recovered for any $\theta$ and $S$. Essentially, this finite-state algorithm uses $S$ states to estimate the number of ones in a sliding window of length $S-1$, thus it is commonly referred to in the literature as "Imaginary Sliding Window" (See Figure~\ref{fig:samaniego}). In this construction, if the machine is in state $i$ and the fresh sample is "$1$", we move to state $i+1$ with probability $1-\frac{i}{S-1}$ or stay in state $i$ with probability $\frac{i}{S-1}$. If the fresh sample is "$0$", we move to state $i-1$ with probability $\frac{i}{S-1}$ or stay in state $i$ with probability $1-\frac{i}{S-1}$. The estimate function is $\hat{\theta}(i)=\frac{i-1}{S-1}$ for $i\in[S]$. 
\begin{center}
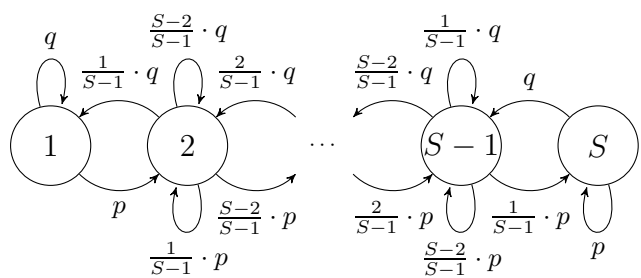
\begin{figure}[H]
\centering
\setlength\belowcaptionskip{-1.4\baselineskip}
\begin{tikzpicture}
  \tikzset{
    >=stealth',
    node distance=1.8cm,
    state/.style={font=\scriptsize,circle, align=center,draw,minimum size=30pt},
    dots/.style={state,draw=none}, edge/.style={->},
  }
  \node [state ,label=center:$1$] (1) {};
  \node [state ,label=center:$2$] (2)   [right of = 1] {};
  \node [dots]  (dots)  [right of = 2] {$\cdots$};
  \node [state,label=center:$S-1$] (S-1) [right of = dots]  {};
  \node [state,label=center:$S$] (S) [right of = S-1]  {};
  \path [->,draw,thin,font=\footnotesize]  (1) edge[loop above] node[above] {$q$} (1);
  \path [->,draw,thin,font=\footnotesize]  (1) edge[bend right=45] node[below] {$p$} (2);
  \path [->,draw,thin,font=\footnotesize]  (2) edge[bend right=45] node[above] {$\frac{1}{S-1}\cdot q$} (1);
  \path [->,draw,thin,font=\footnotesize]  (2) edge[loop above] node[above] {$\frac{S-2}{S-1}\cdot q$} (2);
  \path [->,draw,thin,font=\footnotesize]  (2) edge[bend right=45] node[below] {$\frac{S-2}{S-1}\cdot p$} (dots);
  \path [->,draw,thin,font=\footnotesize]  (2) edge[loop below] node[below] {$\frac{1}{S-1}\cdot p$} (2);
  \path [->,draw,thin,font=\footnotesize]  (dots) edge[bend right=45] node[above] {$\frac{2}{S-1}\cdot q$} (2);
  \path [->,draw,thin,font=\footnotesize]  (dots) edge[bend right=45] node[below] {$\frac{2}{S-1}\cdot p$} (S-1);
  \path [->,draw,thin,font=\footnotesize]  (S-1) edge[bend right=45] node[above] {$\frac{S-2}{S-1}\cdot q$} (dots);
  \path [->,draw,thin,font=\footnotesize]  (S-1) edge[bend right=45] node[below] {$\frac{1}{S-1}\cdot p$} (S);
  \path [->,draw,thin,font=\footnotesize]  (S-1) edge[loop below] node[below] {$\frac{S-2}{S-1}\cdot p$} (S-1);
  \path [->,draw,thin,font=\footnotesize]  (S) edge[bend right=45] node[above] {$q$} (S-1);
  \path [->,draw,thin,font=\footnotesize]  (S-1) edge[loop above] node[above] {$\frac{1}{S-1}\cdot q$} (S-1);
  \path [->,draw,thin,font=\footnotesize]  (S) edge[loop below] node[below] {$p$} (S);
\end{tikzpicture}
\caption{The bias estimation machine of~\cite{samaniego1973estimating} for $\Bern(p)$ input and $q=1-p$}\label{fig:samaniego}
\end{figure}
\end{center}

An analysis of the stationary distribution of the chain proves that this construction asymptotically induces a $\mathrm{Binomial}(S-1,\theta)$ stationary distribution on the memory state space, thus achieving  
\begin{align}
   \E(\hat{\theta}(i)-\theta)^2=\Var\left(\frac{\mathrm{Binomial}(S-1,\theta)}{S-1}\right)=\frac{\theta(1-\theta)}{S-1}. 
\end{align}



\subsubsection*{Leighton and Rivest~\cite{leighton1986estimating} - Lower bound}
The key ingredient in the proof is the use of the \textit{Markov Chain Tree Theorem} (see the elegant proof of Anantharam and Tsoucas~\cite{anantharam1989proof}), which implies that the limiting distribution $\pi_j$ of an $S$-state ergodic Markov chain (which is a unique stationary distribution) can be expressed as ratios of $S-1$
degree polynomials with non-negative coefficients.\footnote{The theorem actually holds in the general case, for which the limiting distribution might not exist, with $\pi_j$ replaced by $\bar{p}_{1j}$, where $\bar{p}_{1j}$ is the long-run average probability that the chain will be in state $j$ given that it
started in state $1$. Specifically, it is the $(1,j)$ element of the matrix $\bar{P}=\lim_{n\rightarrow \infty}n^{-1}\sum_{i=1}^nP^{i-1}$, where $P$ is the transition matrix of the chain, and $\bar{P}$ always exists as $P$ is stochastic.}
Specifically, we have 
\begin{align}
\pi_j=\frac{\sum_{i=1}^S a_{ij} \theta^{i-1}(1-\theta)^{S-i}}{\sum_{i=1}^S a_i\theta^{i-1}(1-\theta)^{S-i}}, \label{eq:tree_thm}
\end{align}
where $a_{ij}\geq 0$ for $1\leq i,j\leq S$, and $a_i = \sum_{j=1}^Sa_{ij}$. Consequently, for any machine $f$ and estimate function $\hat{\theta}$, the quadratic risk is
    \begin{align*}
    R_\theta(f,\hat\theta)=\frac{\sum_{j=1}^S\sum_{i=1}^S a_{ij} \theta^{i-1}(1-\theta)^{S-i}\left(\hat{\theta}(j)-\theta\right)^2}{\sum_{i=1}^S a_i\theta^{i-1}(1-\theta)^{S-i}}.
    \end{align*} 
The remainder of the proof essentially
shows that functions of this restricted form cannot accurately predict $\theta$, by leveraging elementary polynomial analysis to show that there must be some $\theta^*$ for which $R_{\theta^*}(f,\hat\theta)=\Omega(1/S)$. Thus the limitations imposed by restricting the class of transition functions dominate the limitations imposed by quantization of the estimates (in contrary to the Gaussian case where quantization provided the main bottleneck).

\subsection{Deterministic Algorithm for Bias Estimation}
We consider now the bias estimation problem with restriction to deterministic machines. Following the results of Section~\ref{sec:hyp}, one might expect a degradation in performance when comparing to optimal randomized machines. In fact, in the same seminal work of~\cite{leighton1986estimating}, the authors further constructed a deterministic $S$-state estimation algorithm by de-randomizing Samaniego's construction, and as a result showed that $R_{\mathsf{det}}^*(S) = O(\log{S}/S)$. They then conjectured that this is indeed the deterministic minimax risk of the problem. A nice interpretation of their conjecture is the naturally appealing claim that an optimal deterministic algorithm can be obtained by derandomizing an optimal random algorithm. In their deterministic algorithm, randomness is extracted from the measurements via a Von Neumann like procedure~\cite{von195113}, augmenting each state with $O(\log(S))$ additional states\footnote{This derandomization is almost tight. To simulate the algorithm, we need to generate $S$ Bernoulli r.v.s with parameters $i/S$ for $i=1,\ldots,S$ for transition probabilities. Assume we use machines with $N_t$ states respectively. The blowup factor $N$ in the number of states is the average of $N_t$, and there must be at least $S/2$ machines with $N_t \leq 2N$. There are at most $(2N)^{4N+1}$ binary input FSM with at most $2N$ states, hence generating $S/2$ different Bernoulli parameters dictates that $(2N)^{4N+1}\geq S/2$, which implies $N=\Omega(\log(S)/\log\log (S))$.}.  After standing unresolved for a couple of decades, this conjecture was recently disproved by Berg et al.~\cite{berg2021deterministic}, who proved that
\begin{align}
    R_{\mathsf{det}}^*(S)  \asymp\frac{1}{S},
\end{align}
which implies that the memory complexity of the problem is unaffected by the restriction to deterministic algorithms.

This result is proved by an explicit construction of a deterministic $S$-state machine that attains quadratic risk of $O(1/S)$ uniformly for all $\theta\in[0,1]$. The high-level idea is to break down the memory-constrained estimation task into a sequence of memory-constrained (composite) binary hypothesis testing sub-problems as in Wagner's~\cite{wagner1972estimation} scheme, but using time invariant algorithms. In each such sub-problem, the goal is to decide whether the true underlying parameter $\theta$ satisfies $\{\theta<q\}$ or $\{\theta>p\}$, for some $0<q<p<1$. Those decisions are then used in order to traverse an induced Markov chain in a way that enables us to accurately estimate $\theta$. 

In particular, the state space is partitioned into $K=O(\sqrt{S})$ disjoint sets denoted by $\mathcal{S}_1,\ldots,\mathcal{S}_K$, where the estimation function value is the same inside each $\mathcal{S}_i$. These sets are referred to as \emph{mini-chains}, where the $i$th mini-chain consists of $N_i$ states, and is designed to solve the composite binary hypothesis testing problem:
\begin{align}
    \mathcal{H}_0:\left\{\theta>\frac{i+1}{K}\right\} \text{  vs.  } \mathcal{H}_1:\left\{\theta<\frac{i}{K}\right\}.
\end{align}
Each mini-chain $\mathcal{S}_i$ is initialized in its entry state $s_i$, and eventually moves to the entry state $s_{i+1}$ of mini-chain $\mathcal{S}_{i+1}$ if it decided in favor of hypothesis $\mathcal{H}_0$, or to the entry state $s_{i-1}$ of mini-chain $\mathcal{S}_{i-1}$ if it decided in favor of hypothesis $\mathcal{H}_1$. The hypothesis testing mini-chains are taken to be the deterministic testers from~\cite{berg2020binary}, as can be seen in Figure~\ref{fig:mini_chain}. 
\begin{center}
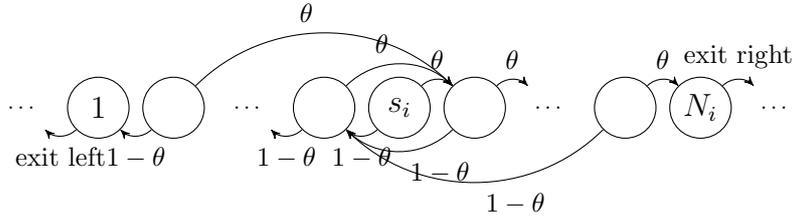
\begin{figure}[H]
\centering
\setlength\belowcaptionskip{-1.4\baselineskip}
\begin{tikzpicture}
  \tikzset{
    >=stealth',
    node distance=0.99cm,
    state/.style={font=\scriptsize,circle, align=center,draw,minimum size=23pt},
    dots/.style={state,draw=none}, edge/.style={->},
  }
  \node [state ,label=center:$1$] (S0)  {} ;
  \node [state] (S0-1)   [right of = S0]   {};
  \node [dots] (dots1)   [right of = S0-1]   {$\cdots$};
  \node [state] (1l) [right of = dots1]  {};
  \node [state ,label=center:$s_i$] (0)   [right of = 1l] {};
  \node [state] (1r) [right of = 0]   {};
  \node [dots]  (dots2)  [right of = 1r] {$\cdots$};
  \node [state] (S1-1) [right of = dots2]  {};
  \node [state ,label=center:$N_i$] (S1) [right of = S1-1]  {};
   \node [dots]  (dots3)  [right of = S1] {$\cdots$};
    \node [dots]  (dots4)  [left of = S0] {$\cdots$};
  \path [->,draw,thin,font=\footnotesize]  (S0-1) edge[bend left=45] node[below ] {$1-\theta$} (S0);
  \path [->,draw,thin,font=\footnotesize]  (0) edge[bend left=45] node[below] {$1-\theta$} (1l);
  \path [->,draw,thin,font=\footnotesize]  (1r) edge[bend left=45] node[below right] {$1-\theta$} (1l);
  \path [->,draw,thin,font=\footnotesize]  (S1-1) edge[bend left=45] node[below right] {$1-\theta$} (1l);
  \path [->,draw,thin,font=\footnotesize]  (1l) edge[bend left=45] node[below ] {$1-\theta$} (dots1);
  \path [->,draw,thin,font=\footnotesize]  (S0-1) edge[bend left=45] node[above left] {$\theta$} (1r);
  \path [->,draw,thin,font=\footnotesize]  (1l) edge[bend left=45] node[above left] {$\theta$} (1r);
  \path [->,draw,thin,font=\footnotesize]  (0) edge[bend left=45] node[above ] {$\theta$} (1r);
  \path [->,draw,thin,font=\footnotesize]  (1r) edge[bend left=45] node[above] {$\theta$} (dots2);
   \path [->,draw,thin,font=\footnotesize]  (S0) edge[bend left=45] node[below] {exit left} (dots4);
  \path [->,draw,thin,font=\footnotesize]  (S1-1) edge[bend left=45] node[above ] {$\theta$} (S1);
 \path [->,draw,thin,font=\footnotesize]  (S1)  edge[bend left=45]  node[above ]{exit right} (dots3);

\end{tikzpicture}
\caption{Mini-chain $\m{S}_i$ for composite binary hypothesis testing}\label{fig:mini_chain}
\end{figure}
\end{center}
Recall that in section~\ref{sec:det_hyp} we saw that $O(K)$ states are needed to resolve between two hypothesis that are $1/K$ apart with constant error probability. Thus, under the assignment $K=O(\sqrt{S})$, the described algorithm can estimate $\theta$ to within $1/\sqrt{S}$ standard deviations using $S$ states, which implies the correct minimax risk.

\subsection{Linear Regression}
\label{subsec:linearregression}

The linear regression problem can be construed as the continuous variant of the parity learning problem. Here the family of distributions $\m{F}$ is characterized by some $\theta\in \mathbb{R}^d$ and is given as a stream of samples $(X_i,Y_i)\in \mathbb{R}^d\times \mathbb{R}$, where the $X_i$'s are drawn from a known i.i.d
distribution $X \sim \m{D}$, and $Y_i=\theta^TX_i+V_i$, where $V_i$ is an i.i.d. noise drawn from a known distribution $\m{D}_V$, independent of $X$ and with zero mean. Our goal is then to estimate the $d$-dimensional vector $\theta$ under the $\ell_2$ loss.  

In~\cite{steinhardt2015minimax}, Steinhardt and Duchi considered the case of memory-bounded sparse linear regression, in which $\theta$ is $k$-sparse, meaning that it has at most $k$ non-zero entries. Essentially, they prove upper and lower bound showing that
\begin{align*}
    \Omega\left(\frac{kd}{b\veps}\right)\leq n^*(2^b,\veps)\leq \tilde{O}\left(\frac{kd}{b\veps^2}\right).
\end{align*}
The lower bound relies on adapting Assouad’s method (see~\cite{yu1997assouad}) to reduce the analysis to bounding the mutual information between a single observation and the parameter
vector $\theta$. The limitation of their proof is in a particular assumption on $X_i$ and letting the additive noise be the unnatural Laplace distribution. They also provide an algorithm that uses $\tilde{O}(b + k)$ bits and requires $\tilde{O}\left(\frac{kd}{b\veps^2}\right)$ observations to achieve error $\veps$ whenever the covariance matrix of $X$ is the identity matrix and $X$ has a small $\ell_{\infty}$ norm. In the upper bound, they show how to implement $\ell_1$-regularized dual averaging using a count sketch data structure, which efficiently counts elements in a data stream. They then use the count sketch structure to maintain a coarse estimate of model parameters, while also exactly storing a small active set of at most $k$ coordinates. These
combined estimates allows to implement dual averaging when the $\ell_1$-regularization is sufficiently large; the necessary amount of $\ell_1$-regularization is inversely proportional to the amount of memory needed by the count sketch structure, leading to a tradeoff between memory and statistical efficiency.

In~\cite{sharan2019memory}, Sharan et al. considered the case of linear regression where the $X$ are drawn independently from a $d$-dimensional isotropic Gaussian and the noise is drawn uniformly from the interval $[-2^{-d/5}, 2^{-d/5}]$. They 
show that any algorithm with at most $d^2/4$ bits of memory requires at least $\Omega(d\log\log 1/\veps)$ samples for $\veps$ sufficiently small as a function of $d$. In contrast, for such $\veps$, $\theta$ can be recovered to error $\veps$ with probability $1-o(1)$ with memory $O(d^2\log (1/\veps))$ bits using $d$ samples. The uniform small error is crucial for the proof, as it reframes the problem as a quantization problem that allows the authors to leverage branching programs for the lower bound. 

\subsection{Additional Results} 
Recently, Jain and Tyagi~\cite{jain2018effective} studied the shrinkage in memory between the hypothesis testing and the estimation problem, namely the interesting fact that a machine with $S$ states can distinguish between two coins with biases that differ by $1/S$, whereas the best additive accuracy it can achieve in estimating the bias is only $1/\sqrt{S}$ (as a result of the previously works of~\cite{hellman1970learning,leighton1986estimating}). The authors explain the memory shrinkage phenomenon by showing that when estimating the unknown bias $\theta$ from an interval using an $S$-state machine, the effective number
of states available is only $O(\sqrt{S})$. Specifically, they show that for any estimator with $S$ states, there exist values $p$ and
$q$ of the bias such that $|p-q|>1/\sqrt{S}$, but
the effective support sets of their equilibrium distributions are
contained in a set of cardinality $O(\sqrt{S})$. In~\cite{kontorovich2012statistical}, Kontorovich defined a consistent estimator as an estimator that converges in probability to $\theta$, and showed that it is impossible to
achieve consistent estimation with bounded memory. However, relaxing the requirement of consistency to $\veps$-consistency, he showed that $O(\log (1/\veps))$ states suffice for a natural class of problems (the binary hypothesis testing problem being one of them, as can also be gleaned from Section~\ref{sec:hyp}).

In the problem of \textit{Universal Prediction} initiated with the works of Merhav and Feder~\cite{merhav1993universal,merhav1993some}, we observe either a worst case or a randomly generated sequence, and are interested in predicting the next value in the sequence. The performance is dominated by the choice of the loss function and is evaluated with respect to an adversary who can see the entire sequence but is limited to the constant predictors family. The figure of merit here is the coding \textit{redundancy}, which is the difference between the optimal regret we can achieve and the one achieved by the adversary. It is known that optimal universal prediction under the log-loss measure is equivalent to optimal encoding of a sequence.

In~\cite{rajwan2000universal}, Rajwan and Feder considered universal finite-memory machines for coding binary sequences, and gave lower bounds for deterministic, randomized, time-invariant and time-varying machines. Meron and Feder~\cite{meron2004optimal} showed that for universal prediction of worst case sequences under the squared error loss, the optimal asymptotic expected redundancy is $O(1/S)$ and it is achieved by the construction of~\cite{leighton1986estimating}. For the log-loss case, using the same construction results in an upper bound on the redundancy of $O(\log S/S)$, while Pinsker's inequality gives a lower bound of $\Omega(1/S)$. In a follow up work~\cite{meron2004finite}, the same authors also consider universal prediction of worst case sequences using deterministic finite-state machines, proving a lower bound on the regret of $\Omega(S^{-4/5})$, and suggested the "Exponentially Decaying Memory" machine, which achieves regret of $O(S^{-2/3})$.

\section{Distribution Property Testing}\label{sec:dist_test}
Distribution property testing is a special case of composite hypothesis testing, where we are interested in deciding whether the underlying distribution has a certain property or is $\veps$-far in total variation from having this property. We focus here on a particular fundamental instance of the property testing problem, known as \textit{uniformity testing}.

In the uniformity testing problem we observe a sequence of independent identically distributed random variables drawn from an unknown distribution $P$ over $[k]$, which is guaranteed to be either the uniform distribution $U_k$, or $\veps$-far from the uniform distribution under the total variation distance. In other words, there are two composite hypotheses in this problem, $\m{F}_0=\{U_k\}$, and $\m{F}_1$ which contains all the distributions $P$ that satisfy $\dtv (P,U_k)> \veps$, and we want to identify the correct (composite) hypothesis with probability of success at least $2/3$. We note that this problem is a particular instance of the identity testing problem, in which $\m{F}_0=\{Q\}$, for some known distribution $Q$ (not necessarily uniform). However, it was shown by Goldreich~\cite{goldreich2016uniform} that the problem of identity testing can be reduced to uniformity testing and, in particular, that for every distribution $Q$ over $[k]$ and every $\veps > 0$, it holds that $\veps$-testing identity to $Q$ reduces to $\veps/3$-testing identity to $U_{6k}$, and that the same reduction can be used for all $\veps > 0$.  

The study of uniformity testing has been initiated by Goldreich and Ron~\cite{goldreich2000testing}, who proposed a simple and natural uniformity tester that relies on the \textit{collision probability} of the unknown distribution, which is the probability that two samples drawn according to $P$ are equal, and requires $O(\sqrt{k}/\veps^4)$ samples. In subsequent work, Paninski~\cite{paninski2008coincidence} proved a lower bound of $\Omega(\sqrt{k}/\veps^2)$ on the sample complexity and provided a matching upper bound that holds under some assumption on $\veps$, which has been later shown to be unnecessary by Diakonikolas et al.~\cite{diakonikolas2014testing}. The lower bound is established by replacing the composite class $\m{F}_1$ with a random distribution, chosen according to the so called \textit{Paninski prior} on $\m{F}_1$. In particular, for any vector $Z\in \{0,1\}^{k/2}$, we define a distribution 
\begin{align}
    P_{\text{pan},i}^{\veps}(Z)= \begin{cases}
    \frac{1+2\veps (-1)^{Z_{i/2}}}{k},& i \text{ even},\\\frac{1-2\veps  (-1)^{Z_{(i+1)/2}}}{k} ,& i \text{ odd}, \end{cases}\label{eq:panin}
\end{align}
for which the value of each coordinate $i\in [k]$ is determined by the vector $Z\in \{0,1\}^{k/2}$. Then, under $\m{H}_1$, we draw $Z\sim \unif\{0,1\}^{k/2}$ and set $P=P_{\text{pan}}^{\veps}(Z)$.
The upper bound is based on collision estimators, since $\sqrt{k}$ samples will result in a single collision on average under the uniform distribution, and $1+\Omega(\veps^2)$ collisions on average under the alternative, and the variances are small enough to distinguish between the hypotheses. 

The memory complexity of estimating the second moment of the empirical distribution (often referred to as the empirical collision probability) is well-studied for worst case data streams of a given length, dating back to the seminal work of Alon et al.~\cite{alon1999space}. In the stochastic setup, Crouch et al.~\cite{crouch2016stochastic} studied the trade-off between sample complexity and memory complexity for the problem of estimating the collision probability. The trade-off between sample complexity and memory / communication complexity for the distribution testing problem has recently been addressed by Diakonikolas et al.~\cite{diakonikolas2019communication}, who gave upper and lower bounds on the sample complexity under the constraint of using at most $b$ memory bits. 

Their upper bound is comprised of a modified uniformity tester that can be implemented in both the memory and communication restricted settings. In particular, a bipartite collision tester is suggested, that works as follows: Draw two independent sets of samples $S_1$ and $S_2$ from the unknown distribution $P$, of sizes $N_1=\frac{b}{2\log k}$ and $N_2=\Theta \left(\frac{k\log k}{b\veps^4}\right)$ respectively, and count the number of pairs of an element of $S_1$ and an element of $S_2$ that collide. Importantly, they have $N_1\cdot N_2\gg k/\veps^4$ which implies $N_1 \ll N_2$, and $\min \{N_1,N_2\} \gg 1/\veps^2$ which implies $b \gg \log k/\veps^2$. The algorithm first stores $S_1$ exactly, and then proceeds to test whether there is some element of $S_1$ that repeats itself more than $10$ times. As the probability of such recurrence is exceedingly small under the uniform distribution, if some symbol in $S_1$ does appear more than $10$ times, the algorithm confidently rejects the uniform hypothesis. This is done to save in the amount of memory needed for the second stage, in which, for each element of the set $S_2$, the algorithm counts the number of collisions this element has with elements of $S_1$. To show that the number of collisions is much larger in the non-uniform case than in the uniform case, note that the expected number of collisions is just $N_1 \cdot N_2 \cdot \lVert P \rVert _2^2$, and by standard concentration bounds one can show that it will likely be close to this value. The amount of memory bits necessary for the algorithm in the first stage is $N_1\cdot \log k$, as we save the entire set $\m{S}_1$. The amount of memory bits necessary in the second stage is at most $\log(10\cdot N_2)$, as the multiplicity of a symbol in $\m{S}_1$ is at most $10$ after the first stage, and that each symbol collides with at most $N_2$ elements of $\m{S}_2$.   
Thus, with at most $N_1\cdot \log k + \log N_2+4< b$ bits and $\Theta \left(\frac{k\log k}{b\veps^4}\right)$ samples, one can solve the uniformity testing problem. 

The lower bound of~\cite{diakonikolas2019communication} is proved in a Bayesian setting using information theoretic tools. Typically, for the purpose of lower bounds, the minimax setting is replaced by a Bayesian one, where the alternative hypothesis is the so-called Paninski prior or \textit{Paniniski mixture}, a uniform mixture over all Paniniski distributions of the form~\eqref{eq:panin} (the specific distribution from the mixture is picked only once, uniformly at random, and then we observe i.i.d. samples from it). A distribution from this mixture is constructed as follows: Let $X=2J-(Z_J\oplus B_J)$, where $J\sim \unif ([k/2]),B\sim \Bern ^{\otimes k/2}(1/2-\veps)$, and $B$ is independent of $J$. Note that for $\veps=0$, this results in the uniform distribution regardless of $Z$. On the other hand, for any choice of $z\in \{0,1\}^{k/2}$, each of these distributions is exactly $P_{\text{pan}}^{\veps}(z)$, which satisfies the total variation condition $\dtv (P_{\text{pan}}^{\veps}(z),U_k)=\veps$, namely is a valid alternative to the uniform hypothesis. 

The lower bound is established for the case where $Z$ is first drawn uniformly on $\{0,1\}^{k/2}$ and then $X_1,\ldots,X_n$ are generated i.i.d. either from $P_{\text{pan}}^{\veps}(Z)$ or from $U_k$. A high-level interpretation of the argument is as follows: For any algorithm, the mutual information between the memory content and $Z$ is at most $b$, so that, from standard rate-distortion results~\cite{cover2012elements}, the best estimate we can get for any $Z_i$ has error $h_2^{-1}(1-\frac{b}{k})=\frac{1}{2}-\Omega\left(\sqrt{\frac{b}{k}}\right)$ on average, where $h_2(\cdot)$ is the binary entropy function, and $h_2^{-1}$ is its inverse restricted to $[0,1/2]$. In each measurement we observe, the random variable $J$ entails no information on whether or not $B_J$ is biased, and hence the informative part of the observation is $Z_J\oplus B_J$, which is unbiased if $B_J\sim\Bern(1/2)$, corresponding to $P=U_K$, and has bias $\Omega\left(\veps\cdot\sqrt{\frac{b}{k}}\right)$ on average, under the alternative hypothesis. Thus, assuming both hypotheses are equally likely a-priori, the mutual information for determining whether the bias of $B_J$ is $0$ or $\veps$, is $O(\veps^2\cdot\frac{b}{k})$ per measurement, and $O(n\veps^2\cdot\frac{b}{k})$ in total.
Thus we need $n=\Omega(k/(b\veps^2))$ samples to succeed. 

Using our definition of the sample-memory curve at $\delta=1/3$, the above results can be written as
\begin{align*}
    \Omega\left(\frac{k}{b\veps^2}\right)\leq n^*(2^b,1/3)\leq \tilde{O}\left(\frac{k}{b\veps^4}\right),
\end{align*}
where the upper bound holds for $b\gg\frac{\log k}{\veps^2}$ bits. A natural question that arises from~\cite{diakonikolas2019communication} is to what extent this memory size can be reduced if we are allowed to process more samples. Another open question posed by the authors is whether the problem can be solved with $o(\log k)$ memory bits. The first question was addressed by Meir~\cite{meir.ITCS.2021}, who showed that one can achieve the sample complexity of~\cite{diakonikolas2019communication} whenever the memory size in bits is $\Omega(\log(k/\veps))$. This is done via comparison graphs, which are a general model for batch collision estimators, whose memory need only store the samples used for collision testing and a global counter.
The second open question was resolved in a recent work of Berg et al.~\cite{berg2022memory} on finding the memory complexity of uniformity testing. In the paper, the authors gave a lower bound of $\Omega (\log k )$ on the memory complexity in bits, thus answering the question from~\cite{diakonikolas2019communication} in the negative. They further proposed an algorithm that succeeds with $\log (k/\veps) +o(\log \log k)$ memory bits. Next, we outline the derivation of these bounds.


\subsubsection*{Upper Bound on the Memory complexity of~\cite{berg2022memory}}
The authors describe an algorithm that succeeds with $O \left(\frac{k \log k}{\veps}\right)$ states by reducing uniformity testing to a sequence of binary hypothesis tests (an approach that should be familiar to us by now). The underlying idea is that if $\dtv(P,U_k) > \veps$, then there must be some symbol $i\in [k]$ such that either $\Pr(X=1 | X \in \{1,i\})$ or $\Pr(X=i | X \in \{1,i\})$ are $\veps$-biased, where by $\veps$-bias we mean equals $1/2+\Omega(\veps)$. On the other hand, if $P=U_k$, none of these probabilities is $\veps$-biased, in fact, they are exactly $\Bern(1/2)$. Following this observation, we can construct a finite state machine whose state space is partitioned into $2k$ disjoint sets (or mini-chains) denoted by $\m{S}_i$, which are identical binary hypothesis testing $\isit(N,1/2+\Omega(\veps),1/2)$ machines with $N= O(1 / \veps \cdot \log k )$ states. These mini-chains test, for all $i>1$, whether $\Pr(X=1 | X \in \{1,i\})$ or $\Pr(X=i | X \in \{1,i\})$ are $\veps$-biased. If some $\m{S}_i$ tests positive for $\veps$-bias, the algorithm decides that $P$ is $\veps$-far from uniform and terminates, otherwise it moves to the next mini-chain, whereas if no test is positive for $\veps$-bias, the machine decides that $P$ is uniform. Due to the properties of the $\isit(N,p,q)$ algorithm, the fact that each machine has $N=O(1 / \veps \cdot \log k )$ states guarantees that Bernoulli parameters that have $O(\veps)$ bias between them can be distinguished with probability of error $\ll 1/k$, thus all tests are successful with some constant error probability ($1/3$, for example). We note that the effective run time of the above algorithm is $O\left(k^2e^{O(1/\veps)}\right)$. This follows since there are $O(k)$ mini-chains that we traverse in sequence, each is activated every $O(k)$ samples on average, and the mixing time of each one is $e^{O(1/\veps)}$.


\subsubsection*{Lower Bound on the Memory complexity of~\cite{berg2022memory}}
The authors show that $\Omega(k+1/\veps)$ states are necessary for uniformity testing. 
For the $\Omega(1/\veps)$ lower bound, they use the fact that a good uniformity tester is also a good binary hypothesis tester
for hypotheses that are $\veps$-far. This is done by choosing the hard setting to be testing between the uniform distribution and a fixed member of the Paninski family~\eqref{eq:panin}. The problem is then reduced to deciding between $\Bern(1/2)$ and $\Bern((1-\veps)/2)$, which, from the lower bound of Hellman and Cover~\cite{hellman1970learning}, must have $\Omega(1/\veps)$ states. 

To obtain the $\Omega(k)$ lower bound, consider first the case of ergodic Markov chains. The proof is based on analysis of stationary distributions and properties of Markov chain transition matrices null space, and it shows that an input distribution can be perturbed in a way that leaves the stationary distribution of the chain unchanged. Thus, there exists a perturbation of the uniform distribution that results in a distribution $\veps$ far from uniform, but induces the same stationary distribution on the state space of the chain. For a given FSM, denote the transition matrix and stationary vector induced by a distribution $p$ as $\Prob(p)$ and $\pi_p$, respectively. We can write the stationary equations under the uniform distribution, $\pi_{U_k}=\pi_{U_k}\Prob({U_k})$, as $\pi_{U_k}=MA{U_k}$, where $M\in [0,1]^{S\times S^2}$ is of the form
\begin{align}
M=\begin{pmatrix}
\pi_{U_k} & 0 & \ldots &0\\
0 & \pi_{U_k} & \ldots & 0\\
\vdots & \vdots & \vdots & \vdots\\
0 & 0 & \ldots & \pi_{U_k}
\end{pmatrix} ,   
\end{align}
and $A\in[0,1]^{S^2 \times k}$ is a matrix whose rows are indexed by $(i,j)\in [S]\times [S]$, such that the entry $A_{(i,j),m}$ specifies the probability of moving to state $j$ from state $i$ when the input is $m\in[k]$.
To simplify, we can write $\pi_{U_k}=MA{U_k}=B{U_k}$, where $B=MA\in [0,1]^{S\times k}$. If $S<k$, the matrix $B$ must have a nontrivial kernel of rank at least $k-S$. Letting $V$ be the subspace spanned by $\ker(B)\cap \ker([1 \ldots 1])$, we have that $\dim(V)\geq k-1-S  \triangleq K$. Using properties of norm contraction, one can show that there exists some vector $x\in V$ with $\lVert x \rVert_\infty=1/k$ and $\lVert x \rVert_1\geq K/k$.
For this $x$, we can define $Q={U_k}-x$.  Recalling that $V$ is orthogonal to $[1 \ldots 1]$, we have that $Q$ is a valid probability distribution and $\dtv{(U_k,Q)}=\frac{1}{2}\lVert x \rVert _1\geq K/2k$. Since $x\in \ker(B)$, we have that $BQ=B{U_k}=\pi_{U_k}$. But $BQ=\pi_{U_k}$ are simply the stationary equations for the chain under $Q$, thus $\pi_Q=\pi_{U_k}$. This implies that for any finite state machine with $S=(1-2\veps)k-1$ states, there exists a distribution $Q$ with $\dtv{(U_k,Q)}\geq \veps$ such that $\pi_{U_k}=\pi_Q$, so the machine cannot distinguish between $U_k$ and $Q$. Thus, an ergodic finite state machine that attains non-trivial error must have $\Omega(k)$ states. 

The $\Omega(k)$ lower bound bound for the non-ergodic case is based on the observation that any $(S,\delta)$ non-ergodic uniformity tester can be reduced to an $(S,\delta +\gamma)$ ergodic uniformity tester (for any $\gamma>0$) by connecting all recurrent states to the initial state with some small positive probability. The result then follows by appealing to the lower bound of the ergodic case above.

\subsection{Additional Results} 
In a recent work, Canonne and Yang~\cite{canonne2024simpler} purposed simpler algorithms for uniformity testing with $n$ samples that use $b$ bits of memory. In their setup, the restriction on the memory size are $b\geq \max \{\log k,\log n\}$ and $b\leq \min\{n\log k,k\log n\}$. The first upper bound follows since straightforward lossless encoding takes $n \log k$ bits, and the second follows since only keeping the number of times each symbol is seen among the $n$ samples, which is a sufficient statistic for the problem, takes $k \log n$ bits. Their first (deterministic) algorithm achieves the same upper bound as~\cite{diakonikolas2019communication}, and relies on a uniformity testing algorithm purposed by Diakonikolas et al.~\cite{diakonikolas2018sample}, which is the total variation distance between the empirical distribution and the uniform distribution. Letting $N_i$ denote the frequency of the symbol $i$, and assuming that $n\leq k$, this statistic is just the normalized number of unseen symbols, that is,
\begin{align}
    Z=\frac{1}{2}\sum_{i=1}^k\left|\frac{N_i}{n}-\frac{1}{k}\right|=\frac{1}{k}\sum_{i=1}^k\ind _{N_i=0}.
\end{align}
Specifically, the algorithm divides the stream of $n$ independent samples into $T$ batches of $s$ samples, which we keep in memory during each batch. At the end of the batch we compute the statistic $Z_t$ and only keep it in memory, while reusing the memory of the $s$ batch samples. This implies $b=\Theta (s\log k)$, and a variance analysis of the estimator shows that the algorithm works as
long as $s^2T=\Omega(k/\veps^4)$. Since $n=s\cdot T$, this results in the upper bound of $n=O\left(\frac{k\log k}{b \veps^4}\right)$.

The second (randomized) algorithm relies on the primitive of \textit{domain compression} introduced by Acharya et al.~\cite{acharya2020inference}, a variant of hashing tailored to distribution testing and
learning, where one can transform testing over domain size $k$ and distance parameter $\veps$ to testing over domain size $k'$ and (smaller) distance parameter $\veps'\asymp\veps\cdot \sqrt{\frac{k'}{k}}$. Letting $k' \asymp b/ \log n$, the memory can now save all the 'new' samples, and due to known results on uniformity testing, the resulting algorithm works as long as the number of samples $n$ satisfies
\begin{align}
    n=O\left(\frac{\sqrt{k'}}{\veps'^2}\right)\Longrightarrow\frac{n}{\sqrt{\log n}}=O\left(\frac{k}{\veps^2\sqrt{b}}\right).
\end{align}

A recent paper by Roy and Vasudev~\cite{roy2023testing} considers distribution testing of a range of properties in the streaming model. They rely on the work of~\cite{diakonikolas2019communication} on uniformity testing
to obtain streaming algorithms for shape-restricted properties. They also consider streaming
distribution testing in other models than the standard i.i.d. sampling one, specifically the
conditional sampling model (see~\cite{chakraborty2013power,canonne2015testing}).

\section{Distribution Property Estimation}\label{sec:prop_est}
The family of distributions is again parameterized by $\m{F}=\{P_{\theta}\}_{\theta \in \mathbb{R}^d}$, but here we are trying to learn some simple scalar function $h(P_{\theta})$ instead of $\theta$ itself. This is usually some continuous (but not one to one) function of the parameter, unlike in the property testing in which we were interested in a discrete function of the parameter. We discuss exclusively the problem of entropy estimation in this section.

\subsection{Entropy Estimation}
In this problem, the parametric family is $\m{F}=\Delta_k$, where $\Delta_k$ denotes the collection of all discrete distributions $P$ over $ [k]$, and we want to estimate the Shannon entropy of $P$, given by $H (P) = -\sum_{x\in [k]}P(x)\log P(x)$, thus $\pi(P)=H(P)$. The loss function is taken to be the $0-1$ loss w.r.t. the $\veps$-error event, and it gives rise to the risk
\begin{align}
	 \Pe (P,\hat{\pi}) = \limsup_{n\rightarrow\infty} \Pr \left(|\hat{\pi}(X^n)-H(P)|>\veps \right)\label{eq:pe_ent}, 
  \end{align}
for $\veps >0$. The minimax risk is thus defined as
\begin{align}
	 \Pe^* (S) = \underset{\hat{\pi} \in \fsm(S)}{\inf}\underset{P\in \Delta_k}{\sup} \Pe (P,\hat{\pi})\label{eq:minimax_ent}. 
  \end{align}
We are generally interested in $\mc(1/3)$, i.e., the smallest number of states needed to guarantee $\veps$-approximate error probability of at most $1/3$ for all $P\in \Delta_k$.

The problem of estimating the entropy of a distribution from i.i.d. samples was addressed by Basharin~\cite{basharin1959statistical}, who suggested the simple and natural \textit{plug-in estimator}, which simply computes the entropy of the empirical distribution of the samples, and requires $\asymp\frac{k}{\veps }+\frac{\log ^2k}{\veps^2}$ samples, thus giving an upper bound on the sample complexity of the problem. In the last two decades, many efforts were made to improve the bounds on the sample complexity of the entropy estimation problem. Paninski~\cite{paninski2004estimating} was the first to prove that it is possible to consistently estimate
the entropy using a sample size sublinear in the alphabet size $k$. The scaling of the sample complexity was shown to be $\frac{k}{\log k}$ in the seminal results of Valiant and Valiant~\cite{valiant2010clt,valiant2011estimating}. The optimal dependence of the sample size on both $k$ and $\veps$ was not completely resolved until recently where, following the works of~\cite{valiant2011power,jiao2015minimax,wu2016minimax}, the sharp sample complexity was shown to be
\begin{align}
  \sc\asymp\frac{k}{\veps \log k}+\frac{\log ^2k}{\veps^2}.  
\end{align} 

The problem of estimating the entropy of a distribution under memory constraints is the subject of a more recent line of work. In~\cite{chien2010space}, Chien et al. addressed the problem of deciding if the entropy of a distribution is above or below some predefined threshold, using algorithms with limited memory. 
The sample-memory curve of entropy estimation was investigated by Acharya et al.~\cite{acharya2019estimating}, who gave an upper bound on the sample complexity under the constraint that the algorithm only uses $O( \log \left(\frac{k}{\veps}\right))$ memory bits, which has the same dependence on $k$ as the plug-in estimator with no memory constraints. Specifically, he constructed an algorithm which is guaranteed to work with $O(k/\veps^3 \cdot \polylog (1/\veps))$ samples and any memory size $b\geq 20 \log \left(\frac{k}{\veps}\right)$ bits (the constant $20$ in the memory size can be improved by a careful analysis, but cannot be reduced to $1$). Their upper bound was later improved by Aliakbarpour et al.~\cite{aliakbarpour2022estimation} to $O(k/\veps^2 \cdot \polylog (1/\veps))$ with memory complexity of $O\left(\log \left(\frac{k}{\veps}\right)\right)$ bits.

In a very recent result~\cite{berg2023entropy}, Berg et al. proved that $\log k+2\log (1/\veps)+o\left(\log k\right)$ bits suffice for entropy estimation when $\veps>10^{-5}$, thus showing that the constant number of $\log k$ bits needed can be reduced all the way down to $1$ (barring that $\veps$ is not too small). Furthermore, the proposed algorithm obtaining the upper bound was shown to converge within $\tilde{O}(k^c)$ samples, for any $c>1$. The algorithm approximates the logarithm of $P(x)$, for a given $x\in[k]$, using a \textit{Morris counter} ~\cite{morris1978counting}, a concept we now briefly explain: Suppose one wishes to implement a counter that counts up to $m$. Maintaining this counter exactly can be accomplished using $\log m$ bits. Morris gave a randomized “approximate counter”, which allows one to retrieve a constant multiplicative approximation to $m$ with high probability using  only $O(\log \log m)$ bits. The Morris Counter was later analyzed in more detail by Flajolet~\cite{flajolet1985approximate}, who showed that $O(\log \log m + \log(1/\veps) + \log(1/\delta))$ bits of memory are sufficient to return a $(1 \pm \veps)$ approximation with success probability $1-\delta$ (this was later improved by Nelson and Yu~\cite{nelson2020optimal}, who showed that $O(\log \log m + \log(1/\veps) + \log\log(1/\delta))$ bits suffice). 

The inherent structure of the Morris counter is particularly suited for constructing a nearly-unbiased estimator for $\log P(x)$, which makes it a judicious choice as a component in memory efficient entropy estimation. The algorithm of~\cite{berg2023entropy} uses two Morris counters, one that approximates a clock, and one that approximates $\log P(X)$.
In order to compute the mean of these estimators, $\E[\widehat{\log P(X)}]$, in a memory efficient manner, a finite-memory bias estimation machine (e.g.,~\cite{leighton1986estimating,berg2021deterministic}) is leveraged for simulating the expectation operator. The bias estimation machine is incremented whenever a count is concluded in the Morris counter approximating the clock, and it results in essentially an average of the estimates over all $x$ values. With $\tilde{O}(k^c)$ samples, this averaging converges (approximately) to the mean of $-\log P(X)$, and thus outputs an approximation to the true underlying entropy.

\section{Conclusions, Outlook, and Open Problems}\label{sec:open}

We discussed the problem of statistical inference under memory constraints and surveyed the state-of-the-art in several problems, including binary hypothesis testing, bias estimation, uniformity testing and entropy estimation. We focused our attention on these particular problems since we see them as canonical problems in statistics, but essentially any other problem in statistical inference can be studied under limited memory constraints. We refrain from specifying particular examples of such open problems, with only one notable exception - the linear regression problem. While this is arguably one of the most extensively studied problems in statistics, current understanding of how this problem is affected by memory constraints remains quite limited. In Section~\ref{subsec:linearregression} we described some known results in this context, and pointed out their limitations. Ideally, one would like to characterize the optimal dependence of the $\ell_2$ loss (or any other loss) on the number of samples $n$, the number of memory states $S$, the dimension $d$ and the noise variance $\sigma^2$, in the setup $Y_i=\theta^T X_i+V_i$, $i=1,\ldots,n$, where $\theta\in\mathbb{R}^d$ belongs to a bounded set or is sparse (or is drawn from some prior), $X_i\stackrel{i.i.d.}{\sim}\m{D}$ for some standard distribution $\m{D}$, say $\m{D}=\m{N}(0,I)$, and $V_i\stackrel{i.i.d.}{\sim}\m{N}(0,\sigma^2)$. First order methods, such as stochastic gradient descent, often give a reasonable tradeoff between the memory and sample cost in this model, but the optimal loss remains unknown (see~\cite{sharan2019memory}).

We have seen that Markov chain theory plays a major role in characterizing the minimax risk in limited memory setting, often in deriving upper bounds by analyzing the induced limiting distributions for certain carefully designed finite-state machines. But this methodology was also used in~\cite{berg2022memory} to obtain a \textit{lower bound} in the uniformity testing problem, by showing that if a machine has a small number of states, then there must exist some distribution that is far from uniform but induces the same stationary distribution on the state space as the uniform distribution. This observation suggests that the problem of statistical inference under memory constraints, in the limit of many samples, is in fact equivalent to question of Markov chain \textit{instability}, namely, to characterizing how sensitive a given chain topology can be to its driving distribution. Let us make this concrete. For a state space with $S$ states and a sample alphabet size $k$, a chain topology is a collection $\m{M} = \{M_s\in \{0,1\}^{k\times S}\}_{s\in[S]}$ of binary matrices, each satisfying $M_s\cdot \mathbf{1} = \mathbf{1}$. We interpret $(M_s)_{i,s'} = 1$ as an edge connecting state $s$ to state $s'$, when the input symbol is $i$. This formulation corresponds to deterministic $S$-state machines, while allowing $M_s\in \{[0,1]^{k\times S}\}_{s\in[S]}$ corresponds to randomized machines. Now, let $P$ be a probability distribution of the samples, represented as a column vector of length $k$. Define the function 
\begin{align}
    \xi_{S,\m{M}}(P)  = \frac{1}{S} \mathbf{1}^T\cdot \lim_{n\to\infty } \left(( I_S \otimes P^T) \cdot \mathsf{diag}\left(\{M_s\}_{s\in [S]}\right)\right)^n, 
\end{align}
whenever the limit exists, where $\otimes$ is the Kronecker product, and $I_S$ is the $S$-dimensional identity matrix. In that case $\xi_{S,\m{M}}(P)$ is the stationary distribution of an $S$-state chain with topology $\m{M}$ under a driving distribution $P$. Now, we can phrase questions about the feasibility of limited-memory inference in terms of $\xi_{S,\m{M}}(P)$. For example, in a simple binary hypothesis testing between $P$ and $Q$, we see that
\begin{align}
    \sup_{\m{M}} \dtv \left(\xi_{\m{S,M}}(P), \xi_{\m{S,M}}(Q)\right) > 1/2
\end{align}
is a necessary condition for $S$ states to suffice. Or, for estimation in a parametric family $\{P_\theta\}$ under some metric loss function $\ell(\hat\theta, \theta)$, we have that 
\begin{align}
    \sup_{\m{M}} \inf_{\ell(\theta,\theta') \geq \epsilon}\dtv \left(\xi_{\m{S,M}}(P_\theta), \xi_{\m{S,M}}(P_{\theta'})\right) > 1/2
\end{align}
is a necessary condition to attain a risk $\epsilon$. Hence, it is interesting to study the joint range 
\begin{align}
\left\{\left(\dtv \left(\xi_{\m{S,M}}(P_\theta), \xi_{\m{S,M}}(P_{\theta'})\right),\ell(\theta,\theta')\right)~:~\theta,\theta'\in\Theta,~\m{M}\right\}    
\end{align}
consisting of all pairs of stationary total variation distance and loss that can be attained by any (deterministic) $S$-state machine.

This survey has focused mainly on characterizing the memory complexity of the problems that were considered, while assuming we have access to an unlimited number of samples. Characterizing the optimal scaling of the risk as a function of both the number of samples $n$ and the number of memory states $S$ is a significantly harder problem, which remains open even for the simplest models one can think of, e.g., the bias estimation model. Information theoretic tools, mainly variants of the information bottleneck concept, were found useful for deriving lower bounds in these scenarios, as described, e.g., in Section~\ref{sec:meth}. The idea is to show that with a small number of memory states, we cannot accurately predict the next sample $X_{t+1}$ from the memory state $U$ that depends only on $X^{t}$ (which implies that we do not learn much about $\theta$ from $U$). To that end, one analyzes the information bottleneck function
\begin{align}
\mathrm{IB}(\log S)=\max_{P_{U|X^t}:I(X^t;U)\leq \log S}I(X_{t+1};U).
\end{align}
In some cases, the restriction to the Markov chain $U-X^t-X_{t+1}$ is too weak to capture the sequential nature of the problem, where $U$ is actually obtained by $t$ consecutive updates. We believe that developing stronger techniques for upper bounding $I(U;X_{t+1})$ is a key step in improving the lower bounds on memory constrained inference.


Finally, we mention a problem in communication with memory constraints, in the same spirit as the problems surveyed in this article, which is worth studying. Consider the problem of communicating over $n$ channel uses of a discrete memoryless channel (DMC) $P_{Y|X}$, when the decoder has limited memory of $S$ states. The question is what is the largest achievable rate. Specifically, an $(n,R,S,\varepsilon)$-code for the DMC $P_{Y|X}$ is specified by an encoder $f:\left[ \lceil 2^{nR} \rceil\right]\to\m{X}^n$, a decoder update function $g:\m{Y}\times [S] \to [S] $ and a decision function $d:\left[S\right]\to \left[ \lceil 2^{nR} \rceil\right]$. For a message $W\sim\mathrm{Uniform} \left[ \lceil 2^{nR} \rceil\right]$ the channel input is $X^n=f(W)$. The channel output $Y^n$ is generated by passing $X^n$ through $P^{\otimes n}_{Y|X}$. The update function is then applied recursively with $U_0=1$ and $U_t=g(Y_t,U_{t-1})$, $t=1,\ldots,n$. Finally, the estimate for $W$ is $\hat{W}=d(U_n)$, and the error probability is $\varepsilon=\Pr(W\neq \hat{W})$. We denote 
\begin{align}
R^*(n,P_{Y|X},S,\varepsilon)&=\max\{R~:~\exists (n,R,S,\varepsilon)-\text{code}\}   \\
C_{\varepsilon}(P_{Y|X},S)&=\sup_{n\in \mathbb{N}} R^*(n,P_{Y|X},S,\varepsilon)
\end{align}
and the goal is to characterize this quantity. Note that for finite output alphabet channels $C_{\varepsilon}(P_{Y|X},S)$ is lower bounded by the finite-blocklength fundamental limit (without memory constraints) $R^*(\log{S}/\log|\m{Y}|,P_{Y|X},\infty,\varepsilon)$, since we can use the $S$ memory states to save $n_S=\log{S}/\log{|\m{Y}|}$ channel outputs. The interesting question here is how much better we can do, if at all, by using $n>n_S$.

\section*{Acknowledgements}
This work was supported by the ISF under Grants 1641/21 and 1766/22.

\bibliography{survey}
\bibliographystyle{ieeetr}
\end{document}